\pdfoutput=1

\documentclass[11pt]{article}

\usepackage[final]{acl}

\usepackage{times}
\usepackage{latexsym}
\usepackage{times}
\usepackage{latexsym}
\usepackage{graphicx}
\usepackage{amsmath}
\usepackage{verbatim}
\usepackage{array}
\usepackage{multirow}
\usepackage{amssymb}
\usepackage{pifont}
\usepackage[skins,breakable]{tcolorbox}
\tcbuselibrary{hooks, breakable}
\usepackage{tikz}
\usetikzlibrary{shapes,arrows}
\usepackage{subcaption}
\usepackage{arydshln} 
\usepackage{graphicx} 
\usepackage{siunitx}
\usepackage{rotating}
\usepackage{tablefootnote}
\usepackage{booktabs}
\usepackage{xcolor}
\usepackage{amssymb}
\usepackage{pifont}
\usepackage{colortbl}
\usepackage[utf8]{inputenc}
\usepackage{csquotes}
\usepackage[utf8]{inputenc}
\usepackage{enumitem}
\usepackage{soul}
\usepackage{algorithm}
\usepackage{algpseudocode}

\usepackage[colorinlistoftodos]{todonotes}

\definecolor{myPastelBlue}{RGB}{153, 219, 237}
\definecolor{myPastelGreen}{RGB}{168, 237, 147}
\definecolor{myPastelRed}{RGB}{242, 136, 136} 
\definecolor{darkgreen}{RGB}{0,128,0}
\definecolor{darkred}{RGB}{128,0,0}
\makeatletter
\tcbset{
  after app={%
    \ifx\tcb@drawcolorbox\tcb@drawcolorbox@breakable
    \else
      \@endparenv
    \fi
  }
}
\usepackage{graphicx} 
\usepackage{adjustbox}
\usepackage{wrapfig}
\usepackage[T1]{fontenc}

\usepackage[T1]{fontenc}

\usepackage[utf8]{inputenc}

\usepackage{microtype}

\usepackage{inconsolata}

\usepackage{graphicx}
\usepackage{tabularx}

\title{MIR: Methodology Inspiration Retrieval for Scientific Research Problems}

\author{
  \begin{tabular}[t]{c}
    Aniketh Garikaparthi$^{1}$ \qquad Manasi Patwardhan$^{1}$ \qquad Aditya Sanjiv Kanade$^{1}$ \\
    Aman Hassan$^{1}$ \qquad Lovekesh Vig$^{1}$ \qquad Arman Cohan$^{2}$
  \end{tabular} \\\\
  $^{1}$TCS Research \quad
  $^{2}$Yale University \\\\
  \small{\texttt{\{aniketh.g, manasi.patwardhan, lovekesh.vig\}@tcs.com}, \quad
  \texttt{arman.cohan@yale.edu}}
}

\begin{document}
\maketitle
\begin{abstract}
There has been a surge of interest in harnessing the reasoning capabilities of Large Language Models (LLMs) to accelerate scientific discovery. While existing approaches rely on grounding the discovery process within the relevant literature, effectiveness varies significantly with the quality and nature of the retrieved literature. We address the challenge of retrieving prior work whose concepts can inspire solutions for a given research problem, a task we define as Methodology Inspiration Retrieval (MIR). We construct a novel dataset tailored for training and evaluating retrievers on MIR, and establish baselines. To address MIR, we build the Methodology Adjacency Graph (MAG); capturing methodological lineage through citation relationships. We leverage MAG to embed an “intuitive prior’’ into dense retrievers for identifying patterns of methodological inspiration beyond superficial semantic similarity. This achieves significant gains of +5.4 in Recall@3 and +7.8 in Mean Average Precision (mAP) over strong baselines. Further, we adapt LLM-based re-ranking strategies to MIR, yielding additional improvements of +4.5 in Recall@3 and +4.8 in mAP. Through extensive ablation studies and qualitative analyses, we exhibit the promise of MIR in enhancing automated scientific discovery and outline avenues for advancing inspiration-driven retrieval.
\end{abstract}

\section{Introduction}

In recent years, Large Language Models (LLMs) have gained popularity as potential tools for novel method synthesis, also referred to as idea generation \cite{Wang_2024, lu2024aiscientistfullyautomated, baek-etal-2025-researchagent, si2024llmsgeneratenovelresearch}. A research problem, either formulated manually \cite{si2024llmsgeneratenovelresearch} or identified from a \textit{seed paper} present in the existing literature \cite{Wang_2024, baek-etal-2025-researchagent}, serves as an input for this task.
\citet{lu2024aiscientistfullyautomated} 
use iterative brainstorming for idea generation with no conditioning on the existing literature. Whereas, \citet{yang2024largelanguagemodelsautomated} condition the process on existing data. This data typically includes knowledge sub-graphs constructed from concepts and relationships of previous studies, citation neighbors of the \textit{seed paper} or semantically similar works 
\cite{Wang_2024, li2024chainideasrevolutionizingresearch, baek-etal-2025-researchagent}. 
This augmentation allows an LLM to ground the generated idea in existing knowledge, reducing the likelihood of hallucinations
\cite{Lewis2020RetrievalAugmentedGF}. However, these approaches face major limitations.

\begin{figure*}[ht]
    \centering  \includegraphics[width=\linewidth]
    {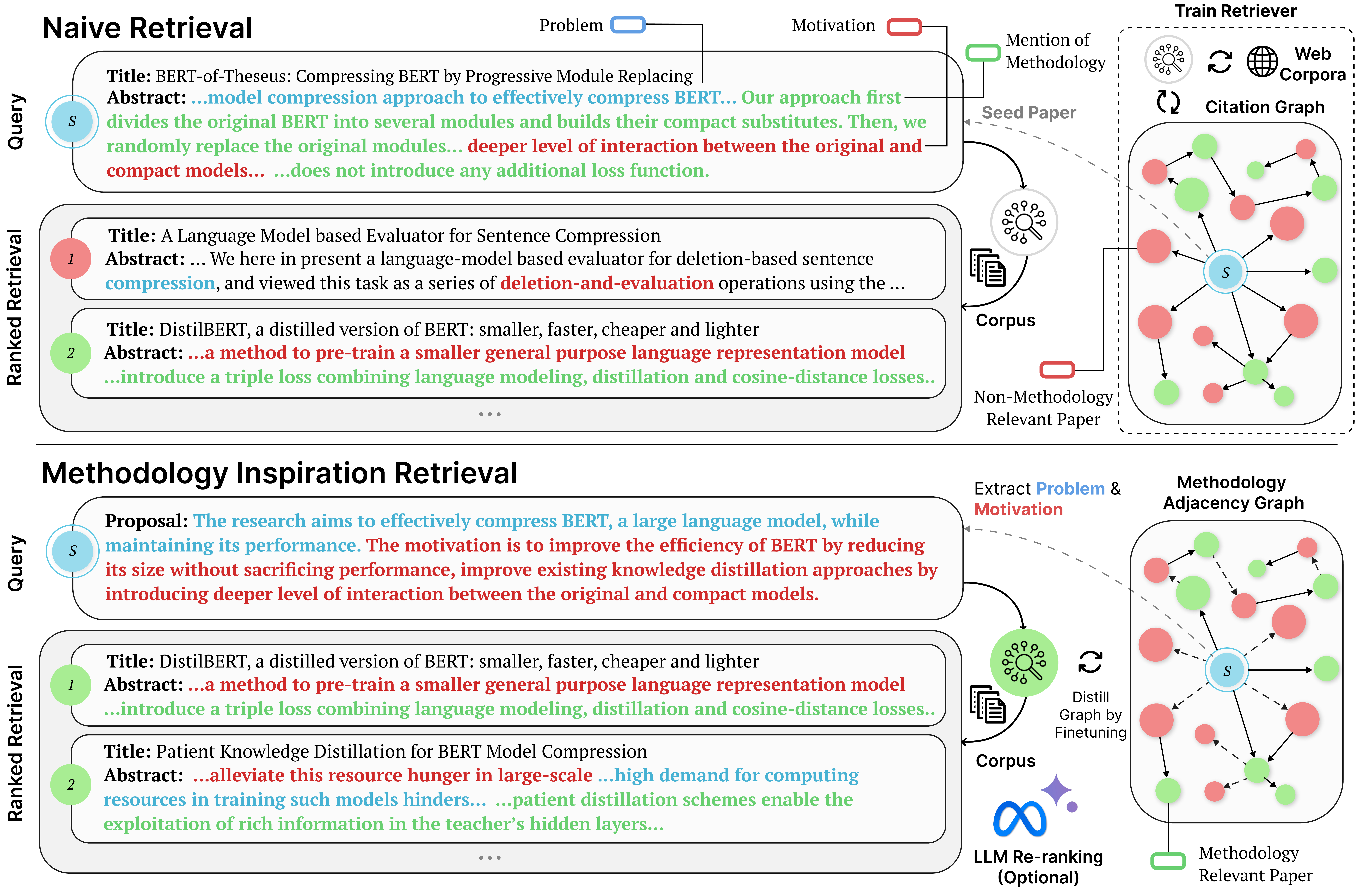}
    \caption{Generic retrieval setting leads to retrieval of non-relevant papers with mere semantic overlap (Top). Whereas, our retriever fine-tuned on a pruned citation graph, learns to surface inspirational literature (Bottom).}
    \label{fig:1}
    \vspace{-0.5em}
\end{figure*}

Firstly, as illustrated in Figure \ref{fig:1} (na\"{i}ve retrieval) the abstracts of the \textit{seed papers}, which are used 
to retrieve and rank relevant literature, have mentions of methodology, experimental design and results of the \textit{seed paper} \cite{li2024chainideasrevolutionizingresearch, baek-etal-2025-researchagent}. 
Having such mentions can bias the retrieval process, reinforcing information from the \textit{seed paper} and thus diminishing the novelty of downstream idea generation. 

Secondly, the retrieval models or APIs\footnote{\href{https://api.semanticscholar.org/}{api.semanticscholar.org}} commonly used to augment papers for idea generation \cite{li2024chainideasrevolutionizingresearch,  Wang_2024, Nigam2024AnIC, pu2024ideasynthiterativeresearchidea, kumar2024largelanguagemodelsunlock, qi2023largelanguagemodelszero}, either employ sentence-based semantic similarity \cite{Reimers2019SentenceBERTSE, Beltagy2019SciBERTAP} or embed generic inter-document relationship 
\cite{Cohan2020SPECTERDR,Singh2022SciRepEvalAM}. This can lead to immaterial retrievals like papers with `keyword overlap', as demonstrated in Figure \ref{fig:1}, where a na\"{i}ve retrieval fetches a paper on \textit{`sentence compression'}, for the problem concerned with \textit{`model compression'}. Whereas, to derive inspiration for developing solutions to a research problem, humans draw from prior corpus not just based on the semantics of the abstract, but through deeper analyses evaluating the potential applicability of different solutions to the investigated problem. For example, as illustrated in Figure \ref{fig:1}, a paper which proposes \textit{`exploitation of rich information in the teacher's hidden layers'}, directly addresses the motivation of \textit{`introducing a deeper level of interaction between the original and compact models'}, and thus serves as a methodological inspiration. 

Lastly, \citet{Wang_2024,guo2024ideabenchbenchmarkinglargelanguage, baek-etal-2025-researchagent} 
assume the availability of citation neighbors of the \textit{seed paper}, presupposing access to the ground truth inspirations, constraining novel methodology generation to be similar to the \textit{seed paper's} methodology. Thus, current approaches depart from the realistic setting, where the key challenge lies in prioritizing retrieval of the literature most conducive to generating novel research ideas, 
provided only the research problem and its motivation as the query. 
Independently, \citet{liu2025researchbenchbenchmarkingllmsscientific} identify \emph{"inspiration retrieval"} as 
one of the 
\textbf{key bottlenecks} towards automated scientific discovery; citing the challenging nature of the task and the importance of deep ``domain intuition'' rather than enhanced reasoning.

To address these gaps, we propose Methodology-Inspiration Retrieval (MIR) (\S\ref{3}), a novel 
task designed to surface methodologically inspirational literature. We envision that a solution for this 
would: (1) benefit automated hypothesis generation as the downstream task and (2) serve as an intelligent recommendation system for researchers seeking inspirations 
to make novel scientific discoveries.

We leverage the rich citation network within a research domain to derive a directed Methodology Adjacency Graph (MAG). 
MAG captures methodological lineage, where edges are annotated with the citation intents pivotal for the task, viz. `methodology' or `non-methodology'.\\
\noindent Our key contributions are: 
\setlist{nolistsep}
\begin{itemize}[noitemsep]
    \item We curate a dataset to fine-tune and evaluate retrievers on MIR by extending the MultiCite dataset \cite{Lauscher_2022}.
    \item  We formulate a novel joint triplet-loss by synthesizing samples from the MAG, lifting Recall@3 by 5.4 and mAP by 7.8 (\S\ref{sec:mir}, \S\ref{6.3}).
    \item  We adapt LLM re-rankers for MIR, adding further gains of 4.5 in Recall@3 and 4.8 in mAP (\S\ref{sec:llm-reranking}, \S\ref{6.3}).
    \item We evaluate the effectiveness of our retrieval on hypothesis generation with LLM-as-a-judge evaluations (\S\ref{6.4}).
\end{itemize}

Finally, we investigate the advantages of our proposed methods, analyze their limitations, and chart actionable directions for advancing inspiration retrieval techniques in future research (\S\ref{6.5}).



\section{Related Works}\label{sec:relwork}

\noindent \textbf{ Hypothesis Discovery:}
Traditionally, hypothesis generation is grounded in the principle of Literature-Based Discovery \cite{04817c26-80f1-38c3-8cf5-03c7fbde591b}, aiming to discover connections between concepts. Recent advances have utilized LLMs to enhance this process by leveraging their comprehension of various scientific domains \cite{Wang_2024, si2024llmsgeneratenovelresearch, baek-etal-2025-researchagent, lu2024aiscientistfullyautomated, li2024mlrcopilotautonomousmachinelearning, Zhou2024HypothesisGW}. Most of these focus on agentic frameworks for generating and refining ideas, but do not prioritize the quality or nature of retrieval for augmentation; limiting the novelty, validity, and diversity of the hypotheses generated \cite{Lewis2020RetrievalAugmentedGF, 
Zhao2024RetrievalAugmentedGF,li2024chainideasrevolutionizingresearch}. In our work, we introduce the task of retrieving papers pertinent to inspiring novel methodologies based solely on a research proposal. 

\textbf{Retrieval of Scientific Documents:}
To retrieve and recommend scientific papers, queries in the form of abstracts and titles \cite{singh-etal-2023-scirepeval, Ostendorff2022NeighborhoodCL, Cohan2020SPECTERDR}, author-specified keywords \cite{SESAGIRIRAAMKUMAR2017577}, detailed textual queries \cite{Anand2017FairScholarBR, parisot-zavrel-2022-multi, medić2023paragraphlevelcitationrecommendationbased}, or specific paper aspects \cite{Singh2024CoSAEmbCS, singh-etal-2023-scirepeval, mysore-etal-2022-multi, Ostendorff2022SpecializedDE}, have been tried in the literature. These approaches learn representations of queries and research papers based on semantic similarity. \citet{yang2024moosechemlargelanguagemodels} also emphasize the ineffectiveness of semantic similarity for scientific retrieval. To address these concerns we attempt to capture an understanding, deeper than semantic relevance, by embedding methodological citation patterns in dense retrievers.


\textbf{Reasoning Intensive Retrieval:}
Information retrieval (IR) has traditionally focused on keyword-based \cite{article} or semantic matching \cite{Devlin2019BERTPO}, overlooking queries that demand reasoning. 
Recent benchmarks (BRIGHT \cite{su2024brightrealisticchallengingbenchmark} and RAR-b \cite{Xiao2024RARbRA}) highlight this gap and the limitations of current approaches. \citet{trivedi-etal-2023-interleaving} 
propose employing Chain-of-thought (CoT) style retrieval, however constructing an effective CoT sequence is non-trivial and the eventual performance hinges upon the \emph{retrieval engine}; essentially performing a task akin to MIR within the CoT chain, thereby underscoring, rather than eliminating the core challenge.
\citet{niu2024judgerankleveraginglargelanguage, yang2024moosechemlargelanguagemodels} employ LLMs as reasoning-intensive re-rankers/retrievers, \citet{Weller2024FollowIREA, Weller2024PromptrieverIR} teach retrievers to follow instructions. LLM-based re-ranking over large corpora is impractical and prohibitively expensive, and IR models are better off without instructions for reasoning tasks \cite{Xiao2024RARbRA}. To address these challenges, we propose 
embedding-based retrieval followed by LLM-based re-ranking, both tailored for the task of MIR. Parallel work \cite{shao2025reasonirtrainingretrieversreasoning} explores contrastive training to adapt retrievers for reasoning-intensive tasks (we detail our differentiation in Appendix \hyperref[B.1]{B.1}).


\section{Problem Definition}
\label{3}
We formally define a new task of \textbf{Methodology Inspiration Retrieval (MIR)}. Specifically, given: 1. A \textbf{Research Proposal}, which consists of a research problem and its motivation $\mathcal{P}$ = \{$\mathcal{R}, \mathcal{M}$\}, for which a novel methodology is to be developed and 2. A \textbf{Literature Corpus} $\mathcal{D}$, consisting of N research papers $\{l^1, l^2, \ldots, l^N\}$. The task is to provide a ranking of top-k papers $\hat{\mathcal{D}_k}$ = $\{\hat{l^{1}}, \hat{l^{2}}, \ldots, \hat{l^{k}}\}$ within the corpus $\mathcal{D}$ based on applicability towards developing a solution for the research problem mentioned in the proposal. We assume that the research proposal $\mathcal{P}$ comes from one of the domains for which papers are present within $\mathcal{D}$. Note that the research proposal does not include any mention of a methodology and hence this is a more realistic and challenging task.

\begin{figure*}[ht]
    \centering
    \includegraphics[width=\linewidth]
    {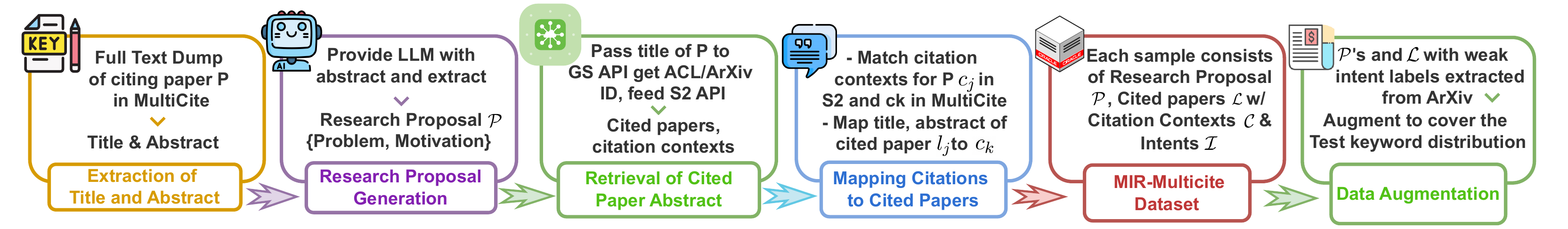}
    \caption{Dataset Construction: Adapting original MultiCite data for MIR and extending it with arXiv augmentation.}
    \label{fig:model_diff}
\end{figure*} \label{fig2}

\section{Dataset Construction} \label{4}
To study this problem, we first curate a dataset for evaluating MIR. For this, we need pairs of research proposals and corresponding research papers which could serve as inspirations to build methodologies for the proposal. We can solicit such pairs from existing citation graphs where the citations have intent labels. 

The MultiCite dataset \cite{Lauscher_2022}, originally designed for citation context intent classification, addresses the key requirements of our task and provides gold-standard citation intent labels.
MultiCite comprises 12,653 citation contexts (one or more sentences) from over 1,200 research papers in the field of computational linguistics in English Language, each annotated with eight distinct citation intents, viz., \textit{Background, Motivation, Future Work, Similar, Difference, Uses, Extension}, and \textit{Unsure} (See Appendix \hyperref[A.2]{A.2} Table \ref{tab:citation_intent_descriptions}). 
We view papers cited with citation intents (\textit{Uses} or \textit{Extension}) as potential 
inspirations. We intentionally maintain this broad definition to capture instances when a cited work offers \emph{conceptual inspiration} beyond immediately applicable solutions. 

This allows us to leverage the MultiCite dataset to construct the \emph{Methodology Adjacency Graph} (MAG) and simulate the MIR task.

\subsection{MIR-MultiCite Dataset}

The multi-step workflow for the construction of MIR-MultiCite is illustrated in \hyperref[fig2]{Figure 2}. We fetch the title and abstract for each citing paper $P$ from MultiCite's full-text dump.
We formulate research proposals $\mathcal{P}$ by extracting the research problems $\mathcal{R}$ and motivations $\mathcal{M}$ from the abstracts of the citing papers using an LLM\footnote{We use Gemini-1.5-pro-001 for all extractions} (Prompt in Appendix \hyperref[A.1.2]{A.1}). We conduct manual analysis on a subset of 30 extracted proposals to evaluate the LLMs' accuracy for this task and find the extractions to be highly accurate (98.2\%), with errors arising from inclusion of traces of methodology in the extracted motivation. To ensure high quality of the test set, we 
correct 2 out of 139 inaccurate extractions.

For each citing paper $P$ with the citation context $c^k$ and intent $i^k$ in Multicite, we need to fetch the cited paper $l^k$. We pass the title and abstract of $P$ to the Google Search API\footnote{\href{https://developers.google.com/custom-search/}{developers.google.com/custom-search}} and/or Semantic Scholar (S2) \cite{Kinney2023TheSS} to determine its ACL or arXiv ID. This is further passed to Semantic Scholar (S2) API to obtain the list of citation contexts and corresponding cited papers $l$ for $P$. For citing paper $P$, if we find a syntactic (subsequence) match between the citation contexts $c^j$ in S2 and $c^k$ in MultiCite, we link cited paper $l^j$ in S2 as $l^k$ with the proposal $\mathcal{P}$, the citation context $c^k$ and intent $i^k$. Otherwise, we extract the bibliography section of $P$ from its full-text dump and feed it along with the citation context $c^k$ including the citation mark to the LLM and task it to determine an entry of the cited paper $l^k$ in the bibliography (Prompt in Appendix \hyperref[A.1.1]{A.1}). We use the title of $l^k$ from the identified entry in the Bibliography to fetch the paper, using the S2 API. We conduct manual verification for 20\% of extractions of $l^k$ for $c_k$ in $P$ and find them to be correct. 

\begin{table}[ht]
\centering
\resizebox{\columnwidth}{!}{
\begin{tabular}{lccccccc}
\toprule
\textbf{Splits} & \textbf{\(\mathcal{P}\)} & \multicolumn{3}{c}{\textbf{Citations}} & \multicolumn{3}{c}{\textbf{Cited Papers}} \\
\cmidrule(lr){3-5} \cmidrule(lr){6-8}
 &  & \textbf{MI} & \textbf{No-MI} & \textbf{Total} & \textbf{MI} & \textbf{No-MI} & \textbf{Total} \\
\midrule
Train       & 704  & 2543  & 5008  & 7551 & 745 & 1124 & 1232 \\
Dev  & 86   & 328    & 676    & 1004 & 95   & 164   & 179 \\
Test        & 139  & 660    & 1698    & 1698 & 193   & 231   & 284 \\
\midrule
\textbf{Total} & \textbf{929} & \textbf{3531} & \textbf{7382} & \textbf{10253} & \textbf{1033} & \textbf{1519} & \textbf{1695} \\
\hdashline
Aug-T & 2216 & 6478 & 16315 & 22793 & 1819 & 4287 & 4688 \\
\bottomrule
\end{tabular}}
\caption{Statistics of the \textit{MIR-MultiCite} dataset. MI: Methodological Intent, Aug-T: Augmented training set}
\label{tab:research-cited-splits}
\end{table}

\vspace{-0.6em}

Our pipeline retains over 80\% of the original MultiCite dataset, forming \textit{MIR-MultiCite}, where each sample has the following structure: 1. A \textbf{Research Proposal}: $\mathcal{P} = \{\mathcal{R}, \mathcal{M}\}$, where $\mathcal{R}$ is the research problem and $\mathcal{M}$ is the motivation behind the problem, 
2. \textbf{Cited Papers: } $\mathcal{L}_{\mathcal{P}} = \{l^1, l^2, \ldots, l^n\}$, where $l^k$ is the $k^{th}$ cited paper in the proposal $\mathcal{P}$, 
3. \textbf{Citation Contexts: } $\mathcal{C}_{\mathcal{P}} = \{ \{c^{1}\}, \{c^{2}\}, \ldots, \{c^{n}\}\}$, where $\{c^{k}\}$ is the set-of citation contexts for the cited paper $l^k$,
\textbf{Citation Intents: } $\mathcal{I}_{\mathcal{P}} = \{\{i^{1}\}, \{i^{2}\}, \ldots, \{i^{n}\}\}$, where $\{i^{k}\}$ is the set-of intents for citation contexts $\{c^{k}\}$. The intent here can be of two types, viz., methodology (\textit{uses}, \textit{extends}) and non-methodology derived from original intent labels in MultiCite. Note that one cited paper $l^k$ can have one or more contexts $\{c^{k}\}$, with distinct intents $\{i^{k}\}$.
These samples formulate a \emph{Methodology Adjacency Graph}, where a node is a Proposal $\mathcal{P}$ with its neighbors as cited papers $\mathcal{L}_{\mathcal{P}}$ and directed edges annotated with intent $\mathcal{I}_{\mathcal{P}}$ and citation contexts $\mathcal{C}_{\mathcal{P}}$. Note that a node is treated as a proposal for an outgoing edge, whereas a cited paper is for an incoming edge. 

Finally, we organize the proposals chronologically. We consider proposals prior to the year 2019 to be part of the train set, from Jan to Jun 2019 to be the dev set, and after Jun 2019 to be the test set. The final dataset statistics are provided in Table \ref{tab:research-cited-splits}.

\subsection{Literature Corpus} \label{4.2}
We consider two settings for simulating the literature corpus $D$:
(i) \emph{Restricted Corpus:} A constrained corpus limited to cited papers within the test set. This introduces potential timeline contamination risks, as there is no guarantee that the corpus papers predate the test proposal. However, it ensures no overlap with the cited papers in the training set.
(ii) \emph{Extended Corpus:} Comprising all cited papers from both the training set and ground-truth citations associated with the test proposals, eliminating timeline contamination. 
More importantly, it tests retriever performance across a more expansive and diverse corpus. Partial contamination has been evidenced in previous literature retrieval works \cite{Cohan2020SPECTERDR}, though not significantly affecting metrics \cite{Ostendorff2022NeighborhoodCL}. However, to maintain transparency we provide results across both settings.

\subsection{Data Augmentation} \label{4.3}
For effective distillation of \emph{MAG}, consistent representation of research subdomains across the splits of \emph{MIR-MultiCite} is crucial. To check the existing dataset's domain representation we employ a state-of-the-art scientific IE system\footnote{\href{https://github.com/thunlp/PL-Marker}{github.com/thunlp/PL-Marker}} and extract keywords from the proposals. We find that the prominent set of keywords for the test set proposals have minimal representation in the training data (Keyword distributions illustrated in Appendix \hyperref[A]{A.2} Figure \ref{fig:keyword_comparison}). To tackle this, we augment our training data using the arXiv Computational Linguistics (cs.CL) corpus\footnote{\href{https://arxiv.org/list/cs.CL/recent}{arxiv.org/list/cs.CL/recent}}. 
Using the S2 API, for $\sim$60k citing papers, we retrieve titles, abstracts and citation texts.
We extract research proposals from these papers and employ SciBERT-based \cite{Beltagy2019SciBERTAP} citation intent classifier\footnote{\href{https://huggingface.co/allenai/multicite-multilabel-scibert}{huggingface.co/allenai/multicite-multilabel-scibert}} trained on the MultiCite, to fetch citation intents as the \emph{weak} labels for the citation contexts. This results in 51,021 proposals amounting to $\sim$2.7 million samples.
We augment only a fraction of these samples to \emph{MIR-MultiCite} train set, allowing consistent domain representation across splits and restricting possible noise introduced by weakly labeled data.
We release the resulting \emph{MIR-MultiCite} dataset at \href{https://github.com/Anikethh/Methodology-Inspiration-Retrieval}{\textcolor{black}{https://github.com/Anikethh/Methodology-Inspiration-Retrieval}}.

\section{Methodology}\label{5}
We first explain how we distill the \emph{MAG} to a retriever by task-specific fine-tuning (\S\ref{sec:mir}, Algorithm \ref{alg:joint-triplet}) and subsequently propose an LLM-based re-ranking strategy tailored for MIR (\S\ref{sec:llm-reranking}).

\subsection{Methodology Inspiration Retriever}\label{sec:mir}
We perform task-specific fine-tuning of existing state-of-the-art retrievers by formulating a triplet loss. We leverage the \emph{MAG} constructed from the training data to synthesize samples for the required triplets. The aim is to bring the representations of a proposal $\mathcal{P}$, closer to a cited paper $l^k$ specifically serving as the methodology inspiration to the problem in $\mathcal{P}$. We utilize the citation context $c^k$ of $l^k$, describing how $l^k$ acts as the methodology inspiration, as an additional signal to serve as a vital link between the $l^k$ and $\mathcal{P}$. This formulation allows the retriever to capture the subtle but significant differences in methodology and non-methodology-related citations for a problem, as opposed to superficial similarities.

With the fine-tuned retriever, we pre-compute the representations of the papers in the literature corpus $D$. During inference, the model only requires a test proposal $\mathcal{P}$ as an input, without needing any information of its cited papers or citation contexts, to produce its embeddings to retrieve the methodologically relevant papers from $D$. 

\subsubsection{Methodology Adjacency Triplet Loss}\label{5.1.1}
We use the following triplet margin loss:
\vspace{-1.1em}
\begin{equation}
L(t, f_\theta) = \max \left\{d(a, p^+) - d(a, p^-)+m,\; 0 \right\}
\label{eq:triplet_loss}
\end{equation}
where $d$ is a distance function and $m$ is the loss margin hyper-parameter.
We use the L2 norm distance:
$d(P_A, P_B) = \| v_A - v_B \|_2  $
where $v_A$ and $v_B$ represent the vectors between which the Euclidean distance is to be calculated. These vectors are computed using a dense retriever $f_\theta$, such that $v = f_\theta(P)$ for any textual input $P$. We use a \textbf{joint triplet loss objective} incorporating three distinct triplets, formulated using samples depicting the relationship between a tuple:  ($\mathcal{P}$, $l^k$, $c^k$). 
We ensure that the triplets synthesized for a sample are in a single batch so that each sample is optimized across all three relationships simultaneously, enabling the model to learn a more cohesive understanding of the methodologically relevant citation patterns.

\subsubsection{Formation of Triplets}\label{5.1.2}
Each triplet consists of an anchor ($a$), a positive sample ($p^+$), and a hard ($p^{--}$) or a soft negative sample ($p^-$):
\[
\textbf{Triplet 1:} 
\tag{2}\label{eq:triplet1}
\begin{cases} 
a = \mathcal{P} \\
p^+ \sim l^k \in \mathcal{L}_{\mathcal{P}} \mid \exists  c^k_j : i^k_j \in \text{\{MI\}} \\ 
p^{--} = l \mid \arg\max\limits_{l \in \mathcal{D} - \mathcal{L}_{\mathcal{P}}} \textit{SIM}( \vec{\mathcal{P}}, \vec{l} ) \ \text{OR} \\
p^- \sim l \in \mathcal{D} - \mathcal{L}_{\mathcal{P}}
\end{cases}
\]
Here, \{MI\} denotes the set-of Methodology Intents, viz. \{`uses', `extends'\}, $\vec{\mathcal{P}}$ and $\vec{l}$ are domain representations and \textit{SIM} is a function computing vector similarity. \textbf{Triplet 1} brings the positively sampled methodologically relevant papers $l^k$ closer to $\mathcal{P}$ relative to the papers that belong to the same domain but may not be methodologically relevant.
To compute `domain' similarity, we extract scientific named entities from $\mathcal{P}$ and abstracts of papers in $\mathcal{D}$ using \cite{Ye2021PackedLM}, categorized into task, method, metric, material, and generic. We store these entities in a text-concatenation format, sorted alphabetically, and create a vector representation using SciBERT \cite{Beltagy2019SciBERTAP}. Note that while we include the named entity \textit{method}, it only serves as a feature for domain representation and may not indicate a problem-methodology relationship
(Appendix \hyperref[B.2]{B.2}). We use this condensed keyword representation to capture the domain of $\mathcal{P}$ and the citing papers and compute `domain' similarity. 
\[
\textbf{Triplet 2:} 
\begin{cases} 
a = c^k \mid i^k \in \{\text{MI}\} \\
p^+ = \mathcal{P} \\
p^{--} \sim \mathcal{P}' \mid  \mathcal{L}' \cap  \mathcal{L} \neq \phi \text{ OR}\\
p^- \sim \mathcal{P}' \mid  \mathcal{P}' \in \mathcal{Q} - \mathcal{P}
\end{cases} \hspace{50pt}
\tag{3}\label{eq:triplet2}
\]
Here $Q$ represents the set of all proposals. \textbf{Triplet 2} brings the representation of 
($\mathcal{P}$) closer to the representation of its solution described through its methodology citation context ($c^k$) rather than the representation of some other
($\mathcal{P}'$) citing similar papers but with a distinct solution. 
\[
\textbf{Triplet 3:} 
\begin{cases} 
a = c^k \mid i^k \in \{\text{MI}\} \\
p^+ = l^k \\
p^{--} \sim l^j \in \mathcal{L} \mid \forall  c^j : i^j \notin \text{\{MI\}} \text{ OR}\\ 
p^-  \sim l \in \mathcal{D} - \mathcal{L}
\end{cases} \hspace{25pt}
\tag{4}\label{eq:triplet3}
\] 
\textbf{Triplet 3} facilitates bringing the representations of the cited paper $l^k$ serving as the inspiration for 
($\mathcal{P}$), closer to the embedding of the citation context $c^k$.
The intent is that these representations are brought closer as opposed to the representations of the cited papers with no methodological applicability. 
We employ a hyper-parameter $\alpha$ for optimal mixing of hard and soft negatives for each triplet type. 

We term our strategy of defining and selecting between $p^{--}$/$p^{-}$ as $\mathcal{S}$: \emph{MAG Guided Sampling}. 


\vspace{-0.5em}
\begin{algorithm}[H]
\caption{Joint Triplet Loss Training Procedure}
\label{alg:joint-triplet}
\begin{algorithmic}[1]
\State \textbf{Input:} Dataset of samples $(\mathcal{P},\mathcal{L}_{\mathcal{P}},\mathcal{C}_{\mathcal{P}},\mathcal{I}_{\mathcal{P}})$, dense retriever $f_\theta$, epochs $E$, batch size $B$, negative sampling ratio $\alpha$
\For{$e=1,\dots,E$}
    \For{each mini-batch of $B$ samples}
        \State Initialize batch loss $\mathcal{L} \leftarrow 0$
        \For{each sample $(\mathcal{P}, \mathcal{L}_{\mathcal{P}}, \mathcal{C}_{\mathcal{P}}, \mathcal{I}_{\mathcal{P}})$}
            \State Select positive citation:
            \State \quad $l^{+}\!\in\!\mathcal{L}_{\mathcal{P}}$ with context $c^{+}\!\in\!\mathcal{C}_{\mathcal{P}}$, \Statex \qquad \hspace{3.1em}    intent$(c^{+})\!=\!\text{MI}$

            \State Sample negative type: $r \sim U(0,1)$
            \State \quad \textbf{if} $r \leq \alpha$: Negative = Hard ($p^{--}$)
            \State \quad \textbf{else}: Negative = Soft ($p^{-}$)

            \State Formulate three triplets $(a,p^+,p^-)$ \Statex \qquad \hspace{2.3em} as per Eqs.~(\ref{eq:triplet1})-(\ref{eq:triplet3}):
            \State \quad $t_1 = (\mathcal{P},\, l^{+},\, l^{-/--})$
            \State \quad $t_2 = (c^{+},\, \mathcal{P},\, \mathcal{P}^{-/--})$
            \State \quad $t_3 = (c^{+},\, l^{+},\, l^{\prime-/--})$

            \State Compute joint loss (Eq.~\ref{eq:triplet_loss}):
            \State \quad $\mathcal{L} \leftarrow \mathcal{L} + L(t_1, f_\theta) + L(t_2, f_\theta) +$
            \Statex \qquad \hspace{9.5em} $L(t_3, f_\theta)$
        \EndFor
        \State $\mathcal{L} \leftarrow \mathcal{L}/B$
        \State Backpropagate $\mathcal{L}$ and update $\theta$
    \EndFor
\EndFor
\end{algorithmic}
\end{algorithm}

\begin{table*}[t]
\centering
\small
\begin{tabular}{l l c *{6}{c}}
\toprule
 & \multicolumn{1}{l}{\textbf{Models}} & \multicolumn{3}{c}{\textbf{Extended Corpus}} & \multicolumn{3}{c}{\textbf{Restricted Corpus}} \\
\cmidrule(lr){3-5} \cmidrule(lr){6-8}
 & \hspace{12em} Metrics $\rightarrow$ & R@3 & R@5 & mAP & R@3 & R@5 & mAP \\
\midrule
\multirow{5}{*}{\rotatebox{90}{\parbox{7em}{\centering \textbf{\textit{Scientific}} \\ \textbf{\textit{Models}}}}} 
& SciBERT ~\cite{Beltagy2019SciBERTAP} & 1.07 & 1.07 & 1.32 & 2.66 & 4.10 & 4.96 \\
& SciNCL ~\cite{Ostendorff_2022} & 37.21 & 44.94 & 35.70 & 53.75 & 62.51 & 50.79 \\
& SPECTER ~\cite{Cohan2020SPECTERDR} & 41.43 & 49.37 & 39.91 & 51.61 & 59.97 & 49.04 \\
& SPECTER2 ~\cite{Singh2022SciRepEvalAM} & 47.21 & 52.75 & 42.41 & 56.61 & 64.72 & 52.45 \\
\cmidrule(lr){2-8}
& SPECTER2 (FT) & 50.33 & 60.35 & 44.38 & 60.21 & 70.07 & 56.89 \\
\midrule
\multirow{10}{*}{\rotatebox{90}{\parbox{10em}{\centering \textbf{\textit{Generic}} \\ \textbf{\textit{Embedding Models}}}}} 
& Qwen-1.5-7B-Inst \cite{yang2024qwen2technicalreport}  & 19.55 & 22.79 & 20.24 & 32.64 & 38.90 & 32.99 \\
& BM-25  \cite{article} & 33.41 & 40.50 & 30.68 & 39.34 & 46.48 & 37.95 \\
& Qwen-2-7B-Inst \cite{yang2024qwen2technicalreport} & 40.08 & 46.36 & 35.50 & 52.33 & 57.30 & 47.45 \\
& ReasonIR \cite{shao2025reasonirtrainingretrieversreasoning} & 51.59 & 59.08 & 45.24 & 60.25 & 64.67 & 53.75 \\
& BGE-EN-ICL \cite{li2024makingtextembeddersfewshot} & 48.30 & 55.94 & 44.54 & 61.40 & 73.38 & 58.02 \\
& Stella\_400M & 49.34 & 55.53 & 48.07 & 64.07 & 70.98 & 59.17 \\
& GritLM-7B \cite{muennighoff2024generativerepresentationalinstructiontuning} & 54.52 & \underline{61.34} & 48.81 & 64.01 & \underline{71.72} & 61.20 \\
& Stella\_1.5B & \underline{54.73} & 60.02 & \underline{49.08} & \underline{64.19} & 71.00 & \underline{63.08} \\
\cmidrule(lr){2-8}
& Stella\_400M (FT) & 59.67 & 64.60 & 56.18 & 66.23 & 72.54 & 63.22  \\
& Stella\_1.5B (FT) & \textbf{60.11} & \textbf{65.32} & \textbf{56.89} & \textbf{66.95} & \textbf{74.70} & \textbf{64.24} \\
\midrule
 & \textbf{Improvements} & \textcolor{darkgreen}{5.38 {$\uparrow$}} & \textcolor{darkgreen}{5.30 {$\uparrow$}} & \textcolor{darkgreen}{7.81 {$\uparrow$}} & \textcolor{darkgreen}{2.76 {$\uparrow$}} & \textcolor{darkgreen}{3.70 {$\uparrow$}} & \textcolor{darkgreen}{1.16 {$\uparrow$}} \\
\bottomrule
\end{tabular}
\caption{Results on the MIR-MultiCite test set. R@k: Recall@k, mAP: mean Average Precision, FT: Task-specific Fine-Tuned. \underline{Underlined}: best-performing base model. \textbf{Bolded}: best-performing model after fine-tuning.}
\label{tab:baselines-comparison}
\end{table*}

\vspace{-1.2em}

\subsection{LLM re-ranking} \label{sec:llm-reranking}
Identifying methodological applicability is often not straightforward, requiring multi-hop reasoning from steps such as identifying sub-problems, establishing analogical links, and recognizing latent relationships. Motivated by the recent success of ~\citet{ sun2023chatgptgoodsearchinvestigating}, we test LLM re-rankers as a complementary approach to first-stage embedding-based retrieval.
We use the dev split proposals and the \emph{elbow} method \cite{Bholowalia2014EBKMeansAC} to find an appropriate $k$ value, such that the top-$k$ retrievals of the fine-tuned retriever cover maximum ground truth methodologically relevant cited papers for the respective proposals. And focus on re-ranking the top-$k$ retrievals.

\subsubsection{Re-ranking Strategies}\label{sec:rankstrat}
We explored several LLM-based re-ranking strategies, including list-wise and pair-wise methods (\hyperref[B.4]{B.4}), but they showed no improvement. Therefore we limit our focus to the following two re-ranking strategies: \textbf{(1) Point-wise:} We provide each of the top-k ranked papers to the LLM and task it to provide a binary relevance judgment, indicating its methodological relevance to the proposal. 
Motivated by \citet{niu2024judgerankleveraginglargelanguage} mimicking the human cognitive processes to perform retrieval with intermediate analyses, we propose \textbf{(2) MIR-Agent:} 
Here we task an LLM to (i) analyze the proposal to list sub-problems and generate a generic action plan to solve the problem, (ii) analyze the applicability of each top-k retrieved paper, 
provided the proposal analysis and (iii) provide a relevance judgment to assess the paper's methodological relevance, given a criteria for assessment along with the outputs from (i) and (ii). We retain the original ordering prioritizing the papers judged to be relevant.

\subsubsection{Re-ranking Settings}\label{sec:rankSet}
We consider four settings with combinations of zero-shot vs. few-shot prompting and using either the abstracts or full text\footnote{We use PyMuPDF \href{https://pymupdf.readthedocs.io/en/latest/}{pymupdf.readthedocs.io/en/latest} to parse and extract text till the end of the \emph{Methodology} section.} of top-$k$ ranked papers. We construct contrastive few-shots from the MAG by sampling methodologically relevant and irrelevant papers for the retrieved most similar proposals from the train set.
Full-paper provides insights into methodologies and discussions often missed with the limited context of the abstract. For this analysis, we identify a sub-set of the test-set proposals for which the full text of top-$k$ $(k = 75)$ retrieved papers is available. 
\section{Results and Discussion} \label{6}

\subsection{Models and hyper-parameter settings} \label{6.1}
We fine-tune retrievers on a single NVIDIA V100 GPU with 32GB of RAM, with a batch size of $4$ for SPECTER2 and Stella 400M, and $1$ for Stella 1.5B and learning rate of $2e^{-6}$. We fine-tune Stella 1.5B with LoRA \cite{hu2022lora}. For LLM re-ranking we use LLaMA-3.1-70B-Instruct \cite{Meta2024TheL3} and Gemini-Pro-001 \cite{Gemini2024} with 32k and 1M context window respecitvely. 

\subsection{Evaluation Metrics} \label{6.2}
We employ \textit{Recall@k} and \textit{mean Average Precision (mAP)} as the evaluation metrics for MIR. Recall@k measures the proportion of ground truth methodologically relevant papers retrieved within the top-k results. For a Proposal, mAP computes the average precision for each ground-truth methodologically relevant cited paper. 

\subsection{Baselines} \label{6.3}
We adopt strong baselines (Table \ref{tab:baselines-comparison}), including scientific retrievers explicitly trained on citation networks, and leading generic retrievers\footnote{\href{https://huggingface.co/dunzhang/stella_en_1.5B_v5}{Stella\_1.5B}, \href{https://huggingface.co/dunzhang/stella_en_400M_v5}{Stella\_400M}} from the MTEB Leaderboard\footnote{\href{https://huggingface.co/spaces/mteb/leaderboard}{huggingface.co/spaces/mteb/leaderboard}} on re-ranking and retrieval tasks. 
We include BM-25 \cite{article} as a strong non-neural baseline. Experiments with retrievers in instruction-following setting are not presented due to poorer performance.

\subsection{Results} \label{6.4}
We discuss the following research questions addressing the effects of our main contributions.

\textbf{RQ1: Does distilling the MAG into the retriever facilitate MIR?}
Both SPERCTER2 and Stella demonstrate consistent improvements over their zero-shot performance after distilling the MAG through fine-tuning (Table \ref{tab:baselines-comparison}), noting up to 10 points gain in metrics for Stella-400M. 
Ablations (Table \ref{tab:ablations}) show a performance drop when fine-tuning without MAG Guided Sampling (\S\ref{5.1.1}), and qualitative analysis in Appendix \hyperref[C.1]{C.1} highlights how MAG-guided sampling improves the ranks of papers with methodological relevance. 

\textbf{RQ2: Does LLM based re-ranking overcome the limitations of embedding-based retrieval for MIR?}
With \( k = 10 \), chosen using the \emph{elbow} method (\S\ref{sec:llm-reranking}), we task LLMs to re-rank the top-k papers based on methodological relevance. We note
improvements of 4.5 in Recall@3 and 4.8 in mAP (Table \ref{tab:llm_reranking}). Through qualitative analysis, we further ablate how (a) few-shots guide the LLMs to better assess
methodological relevance (Appendix \hyperref[C.3]{C.3}, \hyperref[C.4]{C.4}), and (b) MIR-Agent 
provides the intermediate analyses enabling better re-ranking (Appendix \hyperref[C.5]{C.5}). Coupled with a highly effective first-stage retrieval, we limit the resource demands posed by LLMs (Appendix \hyperref[B.6]{B.6}).

\begin{table}[h]
    \centering
    \setlength{\tabcolsep}{4pt}
    \footnotesize
    \begin{tabular}{l|c|c|c|c|c}
        \toprule
        \textbf{Model} & \textbf{Metric} & \multicolumn{2}{c|}{\textbf{Pointwise}} & \multicolumn{2}{c}{\textbf{MIR-Agent}} \\
        \cmidrule(lr){3-4} \cmidrule(lr){5-6}
         &  & ZS & FS & ZS & FS \\
        \midrule
        \multirow{3}{*}{\parbox{1.5cm}{Llama3.1 \\ 70B}} & R@3 $\rightarrow$ & \underline{67.67} & \underline{70.31} & 67.79 & 68.03 \\
         & R@5 $\rightarrow$ & \underline{76.37} & \underline{76.73} & 75.66 & 75.56 \\
         & mAP $\rightarrow$ & \underline{65.62} & \underline{67.05} & 64.65 & 64.73 \\
        \midrule
        \multirow{3}{*}{\parbox{1.5cm}{Gemini1.5 \\ Pro-001}} & R@3 $\rightarrow$ & 67.09 & 69.95 & \underline{71.27} & \underline{\textbf{71.45}} \\
         & R@5 $\rightarrow$ & 72.35 & 75.53 & \underline{\textbf{79.26}} & \underline{79.02} \\
         & mAP $\rightarrow$ & 63.76 & 67.23 & \underline{68.23} & \underline{\textbf{69.02}} \\
        \midrule
        \textbf{\textit{Metric}} & \multicolumn{2}{l|}{\textbf{\textit{Baseline}}} & \multicolumn{2}{c|}{\textbf{\textit{Improvements}}} \\
        \midrule
        R@3 $\rightarrow$ & 66.95 &   \textcolor{darkgreen}{0.72 \textuparrow} & \textcolor{darkgreen}{3.36 \textuparrow} & \textcolor{darkgreen}{4.32 \textuparrow} & \textcolor{darkgreen}{4.50 \textuparrow}\\
        R@5 $\rightarrow$ & 74.70 &   \textcolor{darkgreen}{1.67 \textuparrow} & \textcolor{darkgreen}{2.03 \textuparrow} & \textcolor{darkgreen}{4.56 \textuparrow} & \textcolor{darkgreen}{4.32 \textuparrow}\\
        mAP $\rightarrow$ & 64.24 &   \textcolor{darkgreen}{1.38 \textuparrow} & \textcolor{darkgreen}{2.81 \textuparrow} & \textcolor{darkgreen}{3.99 \textuparrow} & \textcolor{darkgreen}{4.78 \textuparrow}\\
        \bottomrule
    \end{tabular}
    \caption{LLM re-ranking on restricted corpus. Improvements reported for \underline{Underlined} values. Highest values \textbf{Bolded} for each metric. ZS: Zero Shot, FS: Few Shot.}
    \label{tab:llm_reranking}
\end{table}

\textbf{RQ3: Does MIR provide tangible benefits in real-world applications?}
By design, our problem setting aims to prioritize retrieval of papers cited by researchers with methodological intent. Thus, the improvements in retrieval metrics indicate MIR's alignment with real-world user preferences. To better understand the advantages of MIR for the downstream idea generation tasks, we task LLMs to generate ideas grounded with MIR versus na\"{i}ve retrieval. LLM-as-a-Judge idea evaluation (Table \ref{tab:llm-as-a-judge}) evidences higher quality ideas when generations are grounded with MIR, and more aligned with the ground-truth. Details provided in Appendix \hyperref[B.5]{B.5}.

\begin{table}[ht]
\centering
\small
\begin{tabular}{lcc} 
\toprule
\textbf{Metric} & \textbf{Na\"{i}ve Retrieval} & \textbf{MI Retrieval}\\
\midrule
& \textit{IdeaArena} \\
\midrule
Novelty & 996 & 1004 \\
Significance & 985 & 1015 \\
Feasibility & 995 & 1005 \\
Clarity & 992 & 1008 \\
Effectiveness & 981 & 1019 \\
\midrule
\textbf{Overall} & 990 & 1010 \\
\midrule
& \textit{IA-Score} \\
\midrule
Alignment & 0.142 & 0.170 \\
\bottomrule
\end{tabular}
\caption{LLM-as-a-Judge results (top) pairwise evaluation ELO Scores 
\label{tab:llm-as-a-judge}
\cite{li2024chainideasrevolutionizingresearch}, (bottom) alignment with ground-truth methodology \cite{kumar2024largelanguagemodelsunlock}.}
\end{table}

\subsection{Ablations and Error Analysis} \label{6.5}
\textbf{Data Augmentation: } Table \ref{tab:ablations} validates the contribution of data augmentation.
Qualitative analysis in Appendix \hyperref[C.2]{C.2} shows how performance improved for proposals with better keyword representation in the train set. 
Table \ref{tab:stella-train-config-comparison} highlights the robustness and scalability of our training pipeline, showing steady improvements with \emph{only} weakly labeled \emph{MIR-MultiCite-Aug} data.

\begin{table}[ht]
\centering
\tiny
\renewcommand{\arraystretch}{0.9}
\resizebox{\columnwidth}{!}{
\begin{tabular}{l c c}
\toprule
\multicolumn{1}{l}{\textbf{Fine-tuning Setup $\downarrow$}} & \multicolumn{2}{c}{\textbf{mAP}} \\
\cmidrule(lr){2-3}
 & Extended & Restricted \\
\midrule
SPECTER2 (FT) & \textbf{44.38} & \textbf{56.89} \\
w/o $\mathcal{S}$  & 42.83 \textsubscript{\textcolor{red}{(-1.55)}} & 54.77 \textsubscript{\textcolor{red}{(-2.12)}} \\
w/o Aug & 42.17 \textsubscript{\textcolor{red}{(-2.21)}} & 53.00 \textsubscript{\textcolor{red}{(-3.89)}} \\
w/o $\mathcal{S}$ \& Aug & 41.49 \textsubscript{\textcolor{red}{(-2.89)}} & 52.24 \textsubscript{\textcolor{red}{(-4.65)}} \\
\midrule
Stella\_400M (FT) & \textbf{56.18} & \textbf{63.22} \\
w/o $\mathcal{S}$ & 48.89 \textsubscript{\textcolor{red}{(-7.29)}} & 60.70 \textsubscript{\textcolor{red}{(-2.52)}} \\
w/o Aug & 48.57 \textsubscript{\textcolor{red}{(-7.61)}} & 60.67 \textsubscript{\textcolor{red}{(-2.55)}} \\
w/o $\mathcal{S}$ \& Aug & 47.71 \textsubscript{\textcolor{red}{(-8.47)}} & 59.17 \textsubscript{\textcolor{red}{(-4.05)}} \\
\bottomrule
\end{tabular}}
\caption{Ablations. $\mathcal{S}$: MAG Guided Sampling for triplets, Aug: Data augmentation}
\label{tab:ablations}
\end{table}

\begin{table*}[ht]
    \centering
    \setlength{\tabcolsep}{4pt} 
    \footnotesize
    \resizebox{\textwidth}{!}{
    \begin{tabular}{lcccccccc}
        \toprule
        \textbf{Model} & 
        \multicolumn{4}{c}{\textbf{Pointwise}} & 
        \multicolumn{4}{c}{\textbf{MIR-Agent}} \\
        \cmidrule(lr){2-5} \cmidrule(lr){6-9}
         & \multicolumn{2}{c}{Zero Shot} & \multicolumn{2}{c}{Few Shot} & \multicolumn{2}{c}{Zero Shot} & \multicolumn{2}{c}{Few Shot} \\
         \cmidrule(lr){2-3} \cmidrule(lr){4-5} \cmidrule(lr){6-7} \cmidrule(lr){8-9}
         & Abstract & Full Paper & Abstract & Full Paper & Abstract & Full Paper & Abstract & Full Paper \\
        \midrule
        \textbf{Llama3.1 70B} & 
        \shortstack{65.98\textsubscript{R@3}\\70.79\textsubscript{R@5}\\62.23\textsubscript{mAP}} & 
        \shortstack{67.68\textsubscript{R@3}\\71.64\textsubscript{R@5}\\62.90\textsubscript{mAP}} & 
        \shortstack{69.37\textsubscript{R@3}\\72.76\textsubscript{R@5}\\64.01\textsubscript{mAP}} & 
        \shortstack{68.19\textsubscript{R@3}\\72.48\textsubscript{R@5}\\65.27\textsubscript{mAP}} & 
        \shortstack{66.40\textsubscript{R@3}\\75.48\textsubscript{R@5}\\64.84\textsubscript{mAP}} & 
        \shortstack{66.40\textsubscript{R@3}\\74.02\textsubscript{R@5}\\64.79\textsubscript{mAP}} & 
        \shortstack{67.06\textsubscript{R@3}\\74.96\textsubscript{R@5}\\65.11\textsubscript{mAP}} & 
        \shortstack{65.75\textsubscript{R@3}\\74.80\textsubscript{R@5}\\65.04\textsubscript{mAP}} \\
        \midrule
        \textbf{Gemini1.5 Pro-001} & 
        \shortstack{69.92\textsubscript{R@3}\\74.30\textsubscript{R@5}\\65.65\textsubscript{mAP}} & 
        \shortstack{69.37\textsubscript{R@3}\\74.51\textsubscript{R@5}\\66.77\textsubscript{mAP}} & 
        \shortstack{69.88\textsubscript{R@3}\\\textbf{76.21}\textsubscript{R@5}\\64.32\textsubscript{mAP}} & 
        \shortstack{70.49\textsubscript{R@3}\\74.18\textsubscript{R@5}\\65.44\textsubscript{mAP}} & 
        \shortstack{69.94\textsubscript{R@3}\\75.03\textsubscript{R@5}\\65.59\textsubscript{mAP}} & 
        \shortstack{69.66\textsubscript{R@3}\\72.77\textsubscript{R@5}\\65.55\textsubscript{mAP}} & 
        \shortstack{69.24\textsubscript{R@3}\\74.60\textsubscript{R@5}\\66.79\textsubscript{mAP}} & 
        \shortstack{\textbf{72.49}\textsubscript{R@3}\\74.75\textsubscript{R@5}\\ \textbf{68.24}\textsubscript{mAP}} \\
        \midrule
        \multicolumn{3}{l}{\textbf{\textit{Baseline}}} & \multicolumn{2}{c}{\textbf{\textit{Improvements}}} \\
        \midrule\shortstack{66.84\textsubscript{R@3}\\71.64\textsubscript{R@5}\\62.47\textsubscript{mAP}} &  
        \shortstack{\textcolor{darkgreen}{3.08\textsubscript{R@3} \textuparrow} \\ \textcolor{darkgreen}{2.66\textsubscript{R@5} \textuparrow} \\ \textcolor{darkgreen}{3.18\textsubscript{mAP} \textuparrow}} & 
        \shortstack{\textcolor{darkgreen}{2.53\textsubscript{R@3} \textuparrow} \\ \textcolor{darkgreen}{2.87\textsubscript{R@5} \textuparrow} \\ \textcolor{darkgreen}{4.30\textsubscript{mAP} \textuparrow}} & 
        \shortstack{\textcolor{darkgreen}{3.04\textsubscript{R@3} \textuparrow} \\ \textcolor{darkgreen}{4.57\textsubscript{R@5} \textuparrow} \\ \textcolor{darkgreen}{1.85\textsubscript{mAP} \textuparrow}} & 
        \shortstack{\textcolor{darkgreen}{3.65\textsubscript{R@3} \textuparrow} \\ \textcolor{darkgreen}{2.54\textsubscript{R@5} \textuparrow} \\ \textcolor{darkgreen}{2.97\textsubscript{mAP} \textuparrow}} & 
        \shortstack{\textcolor{darkgreen}{3.10\textsubscript{R@3} \textuparrow} \\ \textcolor{darkgreen}{3.39\textsubscript{R@5} \textuparrow} \\ \textcolor{darkgreen}{3.12\textsubscript{mAP} \textuparrow}} & 
        \shortstack{\textcolor{darkgreen}{2.82\textsubscript{R@3} \textuparrow} \\ \textcolor{darkgreen}{1.13\textsubscript{R@5} \textuparrow} \\ \textcolor{darkgreen}{3.08\textsubscript{mAP} \textuparrow}} &
        \shortstack{\textcolor{darkgreen}{2.40\textsubscript{R@3} \textuparrow} \\ \textcolor{darkgreen}{2.96\textsubscript{R@5} \textuparrow} \\ \textcolor{darkgreen}{4.32\textsubscript{mAP} \textuparrow}} &
        \shortstack{\textcolor{darkgreen}{5.65\textsubscript{R@3} \textuparrow} \\ \textcolor{darkgreen}{3.11\textsubscript{R@5} \textuparrow} \\ \textcolor{darkgreen}{5.77\textsubscript{mAP} \textuparrow}}\\
        \bottomrule
    \end{tabular}
    }
    \caption{Full paper re-ranking results with Restricted Corpus. All improvements are reported for Gemini results.}
    \label{tab:full_llm_reranking}
\end{table*}

\begin{table}[ht]
\centering
\footnotesize
\begin{tabular}{l p{0.55cm} *{6}{p{0.60cm}}}
\toprule
\multicolumn{1}{l}{\textbf{Setup}} & \multicolumn{3}{c}{\textbf{Extended Corpus}} & \multicolumn{3}{c}{\textbf{Restricted Corpus}} \\
\cmidrule(lr){2-4} \cmidrule(lr){5-7}
 & R@3 & R@5 & mAP & R@3 & R@5 & mAP \\
\midrule
Base & 49.34 & 55.53 & 48.07 & 64.07 & 70.98 & 59.17 \\
M-Cite & 51.04 & 59.88 & 48.58 & 63.60 & 69.30 & 60.68 \\
arXiv & \underline{54.38} & \underline{61.94} & \underline{52.96} & \underline{64.32} & \underline{71.20} & \underline{62.09} \\
Aug-T & \textbf{59.67} & \textbf{64.60} & \textbf{56.18} & \textbf{66.23} & \textbf{72.54} & \textbf{63.22} \\
\bottomrule
\end{tabular}
\caption{Results on \textit{Stella 400M}. 
M-Cite: training set from MultiCite, arXiv: weakly labeled data from arXiv.}
\label{tab:stella-train-config-comparison}
\end{table}

\textbf{Full Paper in context: } For 75 out of 139 test proposals, with publicly available top-$k$ retrieved full papers, we perform LLM re-ranking providing the full-papers in-context (Table \ref{tab:full_llm_reranking}). We note improvements of 3.25 in recall@3 and 1.45 in mAP in comparison to only providing the abstracts. Appendix \hyperref[C.6]{C.6} shows how the abstract often lacks sufficient context.

\textbf{Triplet Configuration:} Fine-tuning with all triplets instead of a subset performed better across different models and evaluation settings
(Table \ref{tab:stella-comparison}, Appendix \hyperref[B.3.1]{B.3.1}).

\begin{table}[ht]
\centering
\small
\begin{tabular}{l@{\hskip 5pt} p{0.6cm} *{6}{p{0.6cm}}}
\toprule
\multicolumn{1}{l}{\textbf{Config}} & \multicolumn{3}{c}{\textbf{Extended Corpus}} & \multicolumn{3}{c}{\textbf{Restricted Corpus}} \\
\cmidrule(lr){2-4} \cmidrule(lr){5-7}
 & R@3 & R@5 & mAP & R@3 & R@5 & mAP \\
\midrule
Base & 49.34 & 55.53 & 48.07 & 64.07 & 70.98 & 59.17 \\
T1 & 47.54 & 53.38 & 46.22 & 59.04 & 68.08 & 57.04 \\
T1+T2 & 50.80 & 57.63 & 49.17 & 61.44 & 65.23 & 58.73 \\
T1+T3 & 52.12 & 58.71 & 50.39 & 65.16 & \textbf{74.58} & 62.51 \\
T1+T2+T3& \textbf{59.67} & \textbf{64.60} & \textbf{56.18} & \textbf{66.23} & 72.54 & \textbf{63.22} \\
\bottomrule
\end{tabular}

\caption{Results on \textit{Stella 400M}. T1, T2, and T3 refer to Triplets in (\S\ref{5}).}
\label{tab:stella-comparison}
\end{table}

\textbf{Error Analysis:} We analyze 15 proposals with 39 relevant papers, having sub-par Recall@5
. We identify following error types: (1) \textbf{Subjectivity} (7 samples), where papers are methodologically relevant and could be applied, but didn’t align with the research direction adopted by the authors of the seed papers; (2) \textbf{Incorrect/Missing Annotations} (17 samples), where methodologically relevant papers are either mislabeled or not labeled;  and (3) \textbf{Erroneous samples} (15 samples), where our pipeline fails either labeling non-relevant papers as relevant or vice versa (Appendix Section \hyperref[D]{D}). 
We position that further performance improvements would demand finer-grained definitions of methodological relevance indicating the nature and degree of applicability towards research problems.



\section{Conclusion}
Given a research problem as an input, we address the challenge of retrieving papers that can serve as methodological inspirations. We curate a dataset for this task and establish baselines.  Our proposed fine-tuning strategy of sampling triplets from the MAG, augmented with weakly labeled training data helps retrievers develop an implicit domain-intuition for prioritizing inspirational literature. This yields us substantial improvements of +5.4 in Recall@3 and +7.8 in mAP. We further test LLM-based re-ranking as a complementary approach to embedding-based retrieval. Our few-shot guided MIR-Agent strategy with full papers in context yields us improvements of +5.7 in Recall@3 and +5.8 in mAP. We empirically validate the effectiveness of MIR through LLM-as-a-judge evaluations. Finally, our error analysis outlines the avenues for advancing MIR and broader inspiration discovery methodologies.

\section*{Limitations}
\textbf{Domain.} Our dataset is currently restricted to the computational linguistics domain and papers in the English language. However, our domain-specific data augmentation experiment, which leads to substantial improvements in retrieval performance, hints that given the availability of domain-specific research papers, our task-specific fine-tuning and re-ranking techniques can be applicable for the MIR task across domains. \\

\textbf{Documents.} In realistic settings, it is feasible to have only the abstracts of millions of papers in the literature corpus. Hence, we limit fine-tuning our retriever on the abstracts of the cited papers and perform re-ranking of top-k retrieved papers with their full text. Future research could benefit from leveraging long-context retrievers trained directly on full-paper content, potentially capturing richer inspirational signals.\\

\textbf{Annotation.} Our error analysis revealed limitations originating from the original human annotations provided in the MultiCite dataset \cite{Lauscher_2022}. Relevant methodological citations were sometimes omitted due to the original guidelines, which did not require comprehensive annotation of every citation. Additionally, MultiCite’s annotations occasionally missed inspirational citations, while simultaneously including some trivial references (e.g., mere \emph{use} of GPT-4). 
Future work could significantly improve annotation quality and coverage by employing synthetic, LLM-driven citation-intent labeling.\\

\textbf{Subjectivity.} We also observed few ambiguous scenarios where potentially inspirational papers were excluded from the ground-truth set due to differences in the original authors' framing. While this was not a significant bottleneck in our setting, it can compromise the evaluation metric, depending upon the literature corpus and it's distribution. We believe future work should incorporate finer-grained taxonomies to better accommodate diverse inspirational signals.\\

\textbf{Why methodology?} We focus on methodology because, in empirical NLP research, a paper’s central contribution is typically an algorithmic or experimental design, making methodological citations the richest source of actionable inspiration.  In other domains e.g. physics, mathematics, or social sciences, other citation aspects might be more relevant for tracing inspiration as opposed to methodology. However, our proposed pruned citation graph (e.g. MAG) framework, based on mining citation intent patterns, is adaptable and can generalize effectively to other citation aspects (Table \ref{tab:citation_intent_descriptions}). We believe understanding which aspects to consider for other disciplines requires \emph{deep domain understanding} and leave it as part of future work.

We thus frame our contribution as a first step toward principled “inspiration discovery” systems that can help researchers and automated agents map from open problems to novel discoveries.


\bibliography{acl_latex}

\appendix

\onecolumn
\section{Dataset Details} \label{A}
\subsection*{A.1 Extraction Prompts} \label{A.1} 

\begin{tcolorbox}[
    colback=white,
    colframe=pink,
    coltext=black,
    title=\textbf{\textcolor{black}{Extracting Reference}},
    boxrule=0.5mm,
    width=\linewidth,
    arc=2mm,
    boxsep=5pt,
    left=6pt, right=6pt, top=6pt, bottom=6pt,
    breakable
]
Given the following list of references, please provide the title of the paper for the reference: \{reference\_to\_search\} and the citation text\{citation\_text\}\\References:\{references\_text\}\\Your response should only have the title of the paper and no additional comments. If you do not find the correct reference, respond with NOTFOUND. \\Note that the task is extremely crucial and if you have a doubt, respond with NOTFOUND.
\end{tcolorbox} \label{A.1.1}

\begin{tcolorbox}[
    colback=white,
    colframe=pink,
    coltext=black,
    title=\textbf{\textcolor{black}{Extracting Proposal}},
    boxrule=0.5mm,
    width=\linewidth,
    arc=2mm,
    boxsep=5pt,
    left=6pt, right=6pt, top=6pt, bottom=6pt,
    breakable
]
You are an AI assistant tasked with analyzing scientific abstracts. Your goal is to identify three key aspects: Research Problems, Methodologies, and Motivations. Analyze the given abstract and provide your output in JSON format as follows:\\\\
1. Read the abstract carefully.\\
2. Analyze each sentence, considering:\\
- Research Problem (RP): Main issues or challenges being addressed.\\
- Methodology (M): Approaches, techniques, or procedures used.\\
- Motivation (MO): Reasons for defining or solving the research problem.\\
3. For each sentence, provide your thought process on whether it contains RP, M, or MO, and explain why. Note that there can be two aspects as part of the sentence as well.\\
4. After analyzing all sentences, summarize the three key aspects.\\
5. Format your response as a JSON object with the following fields:\\
- "reasoning": Your reasoning about the complete abstract, on how did you analyze and find the research problem, motivation and abstract.\\
- "research\_problem": A fluent summary of the main research problem(s)\\
- "methodology": A coherent summary of the methods and approaches, and results from the paper should not be considered.\\
- "motivation": A clear statement of the motivations behind the research\\\\
Example:\\
Abstract: "We propose a new method for natural language processing. Our approach uses deep learning techniques to improve text classification accuracy. This work is motivated by the increasing need for efficient text analysis in various applications."\\
Output:\\
\{\{
"reasoning": "your reasoning on the abstract",\\
"research\_problem": "The main research problem is improving natural language processing, specifically in the area of text classification.",\\
"methodology": "The researchers propose a new method that utilizes deep learning techniques to enhance text classification accuracy.",\\
"motivation": "The research is driven by the growing demand for efficient text analysis across various applications."\\
\}\}\\\\
Please analyze the given abstract and provide your output in this JSON format.
\end{tcolorbox} \label{A.1.2}

\twocolumn
\subsection*{A.2 Statistics} \label{A.2}
We first provide the descriptions of the citation intents defined in MultiCite, which we borrow for \emph{MIR-MultiCite}.

\begin{table}[ht]
\centering
\begin{tabularx}{\linewidth}{|l|X|}
\hline
\textbf{Citation Intent} & \textbf{Description} \\
\hline
Background & Provides relevant information for this domain.
\\
Motivation &  Provides motivation for the source paper. For instance, it illustrates the need for data, goals, methods, etc. \\
Uses & Uses an idea, method, tool, etc. of the target paper.
 \\
Extends & Builds upon or improves the cited work \\
Similarities & Expresses similarities between the source and the target paper \\
Differences &  Expresses differences towards the target paper or between the source and the target paper\\
Future Work &  Potential avenue for future research. Often corresponds to hedging or
speculative language about work not yet performed.\\
\hline
\end{tabularx}
\caption{Examples of Citation Intents in Scientific Literature}
\label{tab:citation_intent_descriptions}
\end{table} 

\begin{figure}[htbp]
    \centering
    \begin{subfigure}[t]{0.45\textwidth}
        \centering
        \includegraphics[width=\textwidth]{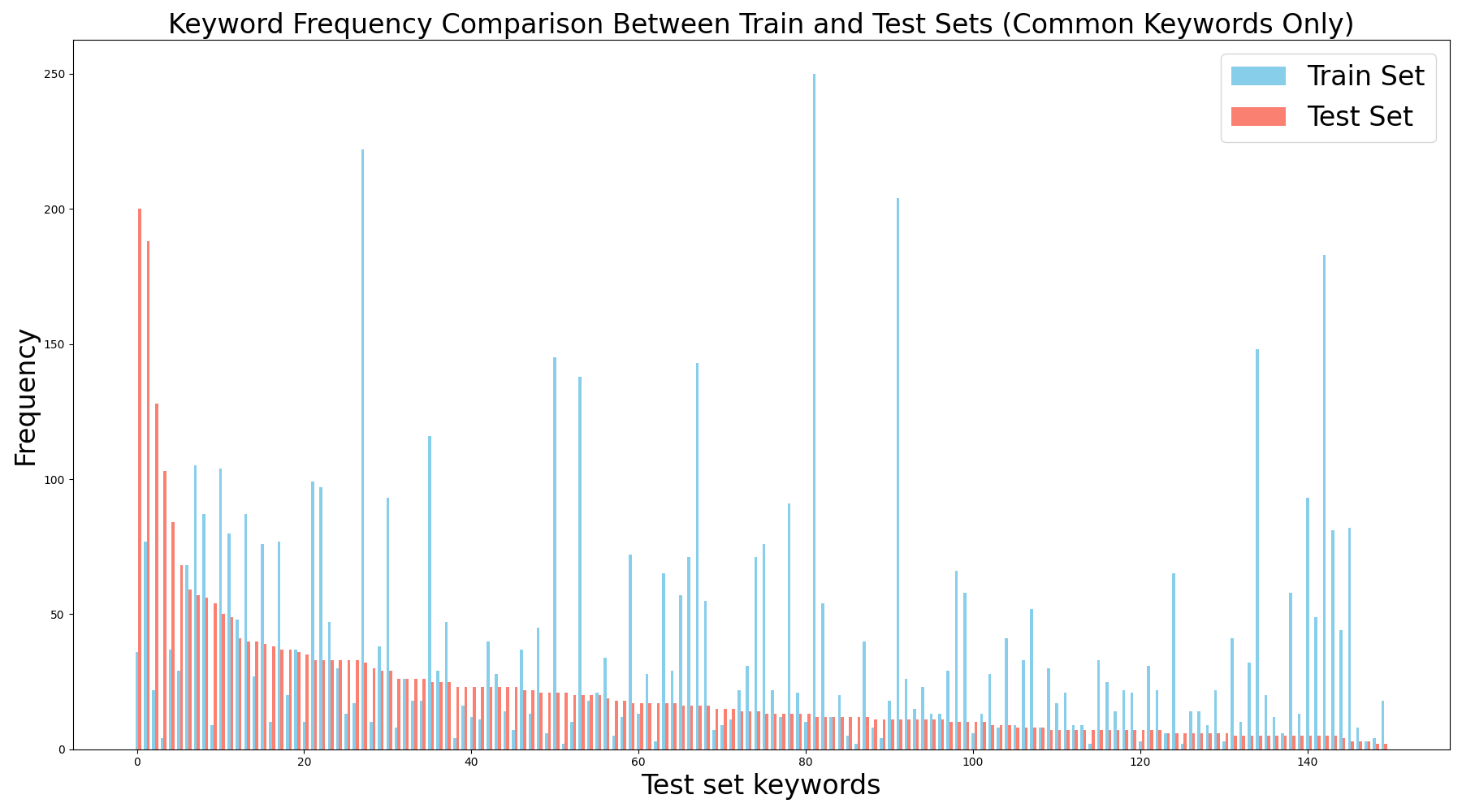}
        \caption{Keyword Frequency in Original Training Data}
        \label{fig:train_keywords}
    \end{subfigure}%
    \hfill
    \begin{subfigure}[t]{0.45\textwidth}
        \centering
        \includegraphics[width=\textwidth]{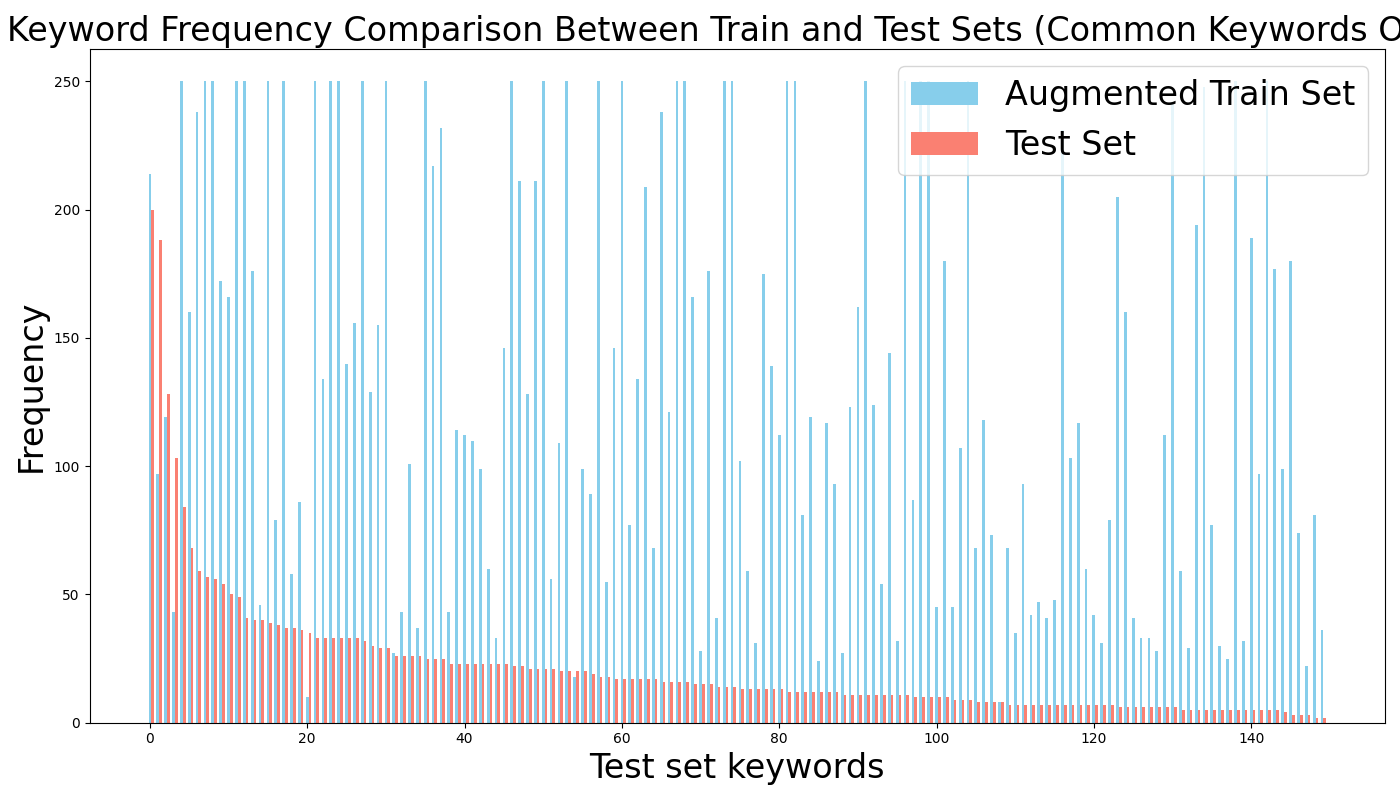}
        \caption{Keyword Frequency after Augmentation}
        \label{fig:aug_train_keywords}
    \end{subfigure}
    \caption{Comparison of Keyword Frequency: Original vs Augmented Training Data}
    \label{fig:keyword_comparison}
\end{figure}

We provide the keyword distributions of train and test sets of \emph{MIR-MultiCite} and \emph{MIR-MultiCite-Aug} in Figures \ref{fig:train_keywords} and \ref{fig:aug_train_keywords} respectively. As it can be seen the high-frequency keywords in the test set are better represented in \emph{MIR-MultiCite-Aug} than \emph{MIR-MultiCite}.

\section{Experimental Settings \& Results}

\subsection*{B.1 Training Objective} \label{B.1}

Our work incorporates a joint triplet-loss learning objective, as this promotes cohesive representation learning (\S\ref{5.1.1}) across multiple aspects; query, document and citation in our case. In comparison, straightforward contrastive learning objectives can miss out on important nuances, that are at the core of scientific publication practices. Citations → [citation texts, citation intents], for example, provide a naturally occurring inter-document incidental supervision signal indicating which documents are most related \cite{Cohan2020SPECTERDR}.

Fine-tuning with only Triplet 1 (Tables \ref{tab:stella-comparison} and \ref{tab:specter2-comparison}), corresponds to training with a simplistic contrastive learning objective, where the representations of positive documents are brought closer, while pushing away representations of negative (and hard negatives) away. 

Our ablations \hyperref[B.3.1]{B.3.1}, confirm that training with only Triplet 1 leads to inconsistent and sub-optimal learning, even decreasing base models performance for Stella-400M.

\emph{Note:} We differentiate only the primary optimization objective from ReasonIR \cite{shao2025reasonirtrainingretrieversreasoning}, and not with their other techniques.


\subsection*{B.2 Domain Representation} \label{B.2}

From the proposal, we extract the following keywords, categorized and sorted:\\
\textbf{Proposal Keywords:} \\ \{BERT compression, model compression, model compression approach, Our approach, GLUE benchmark, knowledge distillation approaches\}  \\\\
We computed similarities with citing paper abstracts, retrieving the Top most similar:  \\\\
\textbf{Highest Similarity Paper 1}\\  
\textbf{Keywords:}
\{context-aware sentiment attention mechanism, knowledge-aware language representation, label-aware masked language model, language representation model, pre-trained language representation models, pre-trained models, language understanding, NLP tasks, pre-training task, SentiLARE, sentiment analysis, sentiment analysis tasks\}  \\\\
\textbf{Highest Similarity Paper 2}\\
\textbf{Keywords:}  
\{data programming approach, data programming paradigm, deep learning architectures, deep learning methods, ML approaches, weak supervision methods, deep learning architectures, discourse analysis literature, expert-composed heuristics, generative model, handcrafted-feature approaches, learning discourse structure, multi-party dialogue, Snorkel framework\}  \\\\
From this sample example, we can see that the use of the \textit{method} entity has a marginal role in computing domain similarity, and empirically it does not introduce methodology-specific sampling biases with potential links between proposal problems and possible solution methods.

\subsection*{B.3 Fine-tuning Ablations}
\subsubsection*{B.3.1 Triplet Configurations} \label{B.3.1}
To validate the need for all three triplets, we provide ablations including three additional fine-tuning setups: (i) Only Triplet 1, (ii) Triplet 1 \& 2, (iii) Triplet 1 \& 3. Table \ref{tab:stella-comparison} shows that (i) decreased performance in comparison to base model, (ii) and (iii) led to sub-optimal improvements and overfit quickly. In contrast using all three triplets led to consistent improvements and better generalization across different models and evaluation settings (Higher performance using subset in Table \ref{tab:specter2-comparison} is likely an anomaly due to data variance, the overall trend supports the effectiveness of all three triplets).

\begin{table}[h]
\centering
\small
\begin{tabular}{l@{\hskip 5pt} p{0.6cm} *{6}{p{0.6cm}}}
\toprule
\multicolumn{1}{l}{\textbf{Config}} & \multicolumn{3}{c}{\textbf{Extended Corpus}} & \multicolumn{3}{c}{\textbf{Restricted Corpus}} \\
\cmidrule(lr){2-4} \cmidrule(lr){5-7}
 & R@3 & R@5 & mAP & R@3 & R@5 & mAP \\
\midrule
Base & 47.21 & 52.75 & 42.41 & 56.61 & 64.72 & 52.45 \\
T1 & 50.23 & 57.97 & \textbf{47.39} & 57.28 & 67.84 & 55.85 \\
T1+T2 & 49.06 & 57.36 & 45.67 & \textbf{60.34} & 69.45 & 54.59 \\
T1+T3 & 48.15 & 56.28 & 46.75 & 58.75 & 66.88 & 54.03 \\
T1+T2+T3 & \textbf{50.33} & \textbf{60.35} & 44.38 & 60.21 & \textbf{70.07} & \textbf{56.89} \\
\bottomrule
\end{tabular}
\caption{Results on \textit{SPECTER2}. T1, T2, and T3 refer to Triplets in (\S\ref{5}).}
\label{tab:specter2-comparison}
\end{table}

\subsubsection*{B.3.2 Performance on Only arXiv Augmented Data} \label{B.3.2}

Table \ref{tab:stella-train-config-comparison}, \ref{tab:specter2-train-config-comparison} demonstrate that our fine-tuning pipeline is robust to potential noise in the training data and performs well even on the arXiv data (weakly labelled) without any human annotated labels. These results highlight the potential of scaling our training recipe for diverse domains through synthetic annotations.

\begin{table}[ht]
\centering
\small
\begin{tabular}{l p{0.55cm} *{6}{p{0.60cm}}}
\toprule
\multicolumn{1}{l}{\textbf{Setup}} & \multicolumn{3}{c}{\textbf{Extended Corpus}} & \multicolumn{3}{c}{\textbf{Restricted Corpus}} \\
\cmidrule(lr){2-4} \cmidrule(lr){5-7}
 & R@3 & R@5 & mAP & R@3 & R@5 & mAP \\
\midrule
Base & 47.21 & 52.75 & 42.41 & 56.61 & 64.72 & 52.45 \\
M-Cite & 41.85 & 55.68 & 41.67 & 56.98 & 68.59 & 52.88 \\
arXiv & \underline{48.44} & \underline{57.96} & \textbf{45.42} & \textbf{62.16} & \textbf{70.98} & \textbf{56.97} \\
Aug-T & \textbf{50.33} & \textbf{60.35} & \underline{44.38} & \underline{60.21} & \underline{70.07} & \underline{56.89} \\
\bottomrule
\end{tabular}
\caption{Results on \textit{SPECTER2}.
M-Cite: training set from MultiCite, arXiv: weakly labeled data from arXiv.}
\label{tab:specter2-train-config-comparison}
\end{table}

\subsection*{B.4 Re-ranking strategies} \label{B.4}
Despite extensive prompt tuning and exploring different settings, the results displayed high variability and an overall drop in performance with both the methods, hence we did not proceed with them for our final results.\\
\textbf{(1) Listwise:} We provide top-k retrieved papers to the LLM and task it to re-rank them based on their methodological relevance to the given proposal \\
\textbf{(2) Pairwise:} We implement a tournament-style pairwise comparison. Each round involves intelligently creating $k/2$ pairs of the retrieved papers from the last round such that similar performing papers are pitched against each o8ther. For each pair, the LLM is tasked to judge which paper shares higher methodological insight to the proposal. We assign a point to the paper with more insight. We aggregate scores across three rounds by summing the wins for each paper to rank them.

\subsubsection*{B.4.1 Re-ranking on Extended Corpus}

Re-ranking on the extended corpus was performed on top-20 papers.

\begin{table}[ht]
    \centering
    \setlength{\tabcolsep}{4pt}
    \footnotesize
    
    \begin{tabular}{l|c|c|c|c|c}
        \toprule
        \textbf{Model} & \textbf{Metric} & \multicolumn{2}{c|}{\textbf{Pointwise}} & \multicolumn{2}{c}{\textbf{MIR-Agent}} \\
        \cmidrule(lr){3-4} \cmidrule(lr){5-6}
         &  & ZS & FS & ZS & FS \\
        \midrule
        \multirow{3}{*}{\parbox{1.5cm}{Llama3.1 \\ 70B}} & R@3 $\rightarrow$ & \underline{60.35} & 62.15 & 60.60 & 61.20 \\
         & R@5 $\rightarrow$ & \underline{65.08} & 67.12 & 65.44 & 67.12 \\
         & mAP $\rightarrow$ & \underline{56.30} & 58.24 & 56.85 & 59.08 \\
        \midrule
        \multirow{3}{*}{\parbox{1.5cm}{Gemini1.5 \\ Pro-001}} & R@3 $\rightarrow$ & 59.40 & \underline{62.25} & \underline{61.50} & \underline{\textbf{63.00}} \\
         & R@5 $\rightarrow$ & 64.26 & \underline{\textbf{67.96}} & \underline{67.59} & \underline{67.38} \\
         & mAP $\rightarrow$ & 55.98 & \underline{58.43} & \underline{59.61} & \underline{\textbf{60.14}} \\
        \midrule
        \textbf{\textit{Metric}} & \multicolumn{2}{l|}{\textbf{\textit{Baseline}}} & \multicolumn{2}{c|}{\textbf{\textit{Improvements}}} \\
        \midrule
        R@3 $\rightarrow$ & 60.11 & \textcolor{darkgreen}{0.24 \textuparrow} & \textcolor{darkgreen}{2.14 \textuparrow} & 
        \textcolor{darkgreen}{1.39 \textuparrow} &
        \textcolor{darkgreen}{2.89 \textuparrow} \\
        R@5 $\rightarrow$ & 65.32 & \textcolor{darkred}{0.24 \textdownarrow} & \textcolor{darkgreen}{2.64 \textuparrow} & \textcolor{darkgreen}{2.27 \textuparrow} & \textcolor{darkgreen}{2.06 \textuparrow} \\
        mAP $\rightarrow$ & 56.89 & \textcolor{darkred}{0.59 \textdownarrow} & \textcolor{darkgreen}{1.54 \textuparrow} & \textcolor{darkgreen}{2.72 \textuparrow} & \textcolor{darkgreen}{3.25 \textuparrow} \\
        \bottomrule
    \end{tabular}
    \caption{LLM re-ranking on extended corpus. Improvements reported for \underline{Underlined} values. Highest values \textbf{Bolded} for each metric. ZS: Zero Shot, FS: Few Shot.}
    \label{tab:llm_ext_reranking}
\end{table}

\subsection*{B.5 LLM-as-a-Judge to assess effectiveness of retrieval on hypothesis generation} \label{B.5}

We evaluate MIR-augmented generations on two methods; \citet{li2024chainideasrevolutionizingresearch}'s IdeaArena evaluating quality of open-ended hypothesis generation through pair-wise comparisons on criteria like novelty, feasibility etc. And \citet{kumar2024largelanguagemodelsunlock}'s IA-Score which measures alignment of LLM generation with ground-truth idea, which we adapt for our setting as the ground-truth methodology.

\subsubsection*{B.5.1 IdeaArena} \label{B.5.1}

For a subset of 30 proposals from our test set, we task the LLM to generate 10 ideas for each proposal employing the template provided by \cite{si2024llmsgeneratenovelresearch}. To reduce any potential bias, we randomly sample 5 papers’ abstracts from the top 20 retrieved papers (MIR vs Na\"{i}ve) and append them in context for generating a novel scientific idea. After generation, we use IdeaArena, as proposed by \cite{li2024chainideasrevolutionizingresearch}, which leverages pairwise comparisons via LLMs to determine ELO scores. We use Gemini-2.0-Flash-Exp to balance cost and performance. 

\subsubsection*{B.5.2 IA-Score} \label{B.5.2}

We employ the following steps to make fair comparisons between MIR and Na\"{i}ve retrieval:

\setlist{nolistsep}
\begin{itemize}[noitemsep]
    \item \textbf{Setup:} For proposals (P) in our test set, we used Gemini-2.0-Flash to generate 5 potential methodology ideas.
    \item \textbf{Grounding Conditions:} The LLM generation was grounded by augmenting its prompt with the top-3 retrieved abstracts obtained via:
    (a) Naïve Retrieval: Using our best baseline retriever
    (b) MIR Retrieval: Using our best fine-tuned MIR retriever
    \item \textbf{Evaluation:} We adapt "Idea Matcher" to evaluate the generated methodologies. An LLM compares each of the 5 generated methodologies against the ground truth methodology (extracted from the original seed paper abstract of P) and assigns a score (1 → match, 0 → no match). The average score across generated ideas for a proposal gives its IAScore.
    \item \textbf{Results:} We averaged IAScores across all test set proposals for each grounding condition:
    \begin{itemize}
        \item \textbf{MIR:} 0.170 (\textit{1 in 6} Generations match GT Methodology)
        \item \textbf{Naïve Retrieval:} 0.142 (\textit{1 in 7} Generations match GT Methodology)
    \end{itemize}
\end{itemize}

This represents a \textbf{20\%} improvement when grounding with MIR. It's important to note that our application targets a task distinct from the original IAScore work: which generated future work ideas from a full seed paper and averaged a score ~0.3 (CS domain). Our setting is inherently more open-ended (justifying the lower score), and not optimized towards alignment with ground-truth methodologies, however we still show improvement over na\"{i}ve retrieval techniques.

\subsection*{B.6 Resource demands of LLM Re-ranking} \label{B.6}

To provide clarity around the resource demands of our LLM re-ranking strategy, we emphasize the effectiveness of our dual-stage; embedding based retrieval followed by LLM re-ranking pipeline, which greatly reduces the resource demands.

We present a cost analysis of Gemini 1.5 Pro for all re-ranking configurations Point-wise/MIR-Agent re-ranking with abstract/full-paper in context.

\begin{table}[ht]
\centering
\setlength{\tabcolsep}{12pt}
\begin{tabular}{|l|c|c|}
\toprule
\multirow{2}{*}{\textbf{Re-ranker}} & \multicolumn{2}{c|}{\textbf{Configuration}} \\
\cline{2-3} 
 & \textbf{Abstract} & \textbf{Full Paper} \\ 
\midrule
Pointwise & \$0.010 & \$0.021 \\
\midrule
MIR-Agent & \$0.023 & \$0.038 \\
\bottomrule
\end{tabular}
\caption{Cost analysis for LLM re-ranking with abstract and full paper for both Point-wise and MIR-Agent re-ranking using Gemini-1.5-Pro-001.} \label{tab:cost_analysis}
\end{table}

The calculations done in Table \ref{tab:cost_analysis} are for a single proposal. In effect a proposal's top-10 retrievals are passed to the LLM for re-ranking.

\onecolumn

\subsection*{B.7 Re-ranking Prompts} \label{tab:full_paper_reranking}
We only position LLMs to output binary relevance judgements to assess methodological applicability. The numerical floating values are only added to aid the LLMs reasoning process and are not used further.
\begin{tcolorbox}[
    colback=white,
    colframe=pink,
    title=\textbf{\textcolor{black}{Pointwise Re-ranking Prompt}},
    boxrule=0.5mm,
    width=\linewidth,
    arc=2mm,
    boxsep=5pt,
    left=6pt, right=6pt, top=6pt, bottom=6pt,
    breakable
]
You are an expert researcher in the field of NLP with rich domain knowledge. You will be provided with a Research Proposal consisting of a problem and motivation, and an \{input\_text\_name\}. \\
Your task is to evaluate the methodological relevance of the Input Paper to the Research Proposal based on the provided criteria. You need to assess whether the methodologies proposed in the \{input\_text\_name\} can contribute towards solving the problem outlined in the Research Proposal. \\

\textbf{Core Assessment Objective} \\
Determine if the input paper offers \textbf{substantive methodological utility} that can be: \\
- Used/Extended/Adapted for solving the problem stated in the Research Proposal. \\

\textbf{Methodological Relevance Criteria} \\

1. \textbf{Problem Proximity} \\
- Addressing \textbf{similar} core research problems. \\
- Potential for adaptable solution approaches. \\
- Providing potentially transferable techniques. \\

2. \textbf{Methodological Utility} \\
- \textbf{Complementarity}: Do the methods fill gaps or enhance the proposal's approach? \\
- \textbf{Generalizability}: Can the methods be extended or adapted to the problem setting of the proposal? \\
- Evaluate whether the methods offer fresh perspectives or synergize with the proposal's needs. \\

Classify as \textbf{Relevant (1)} if: \\
- The problems of both the proposal and input paper are conceptually related. \\
- The input paper offers baselines to build upon or foundational models/architectures/evaluation metrics that could be adapted. \\
- There is potential for a direct knowledge transfer across domains. \\

Classify as \textbf{Non-Relevant (0)} if: \\
- Orthogonal research domains and no discernible problem relationship. \\
- The input paper offers no clear methodological utility. \\
- Prohibitively complex methodology transfer. \\

\textbf{Exemplars} \\
\{exemplars\} \\\\
These are only examples to understand what kind of papers are considered methodologically relevant, and which are not, these are not related to the actual papers you will be evaluating. \\\\
\textbf{Test Proposal}\\
\{proposal\}\\
\textbf{\{input\_text\_name\}} \\
Title: \{input\_paper\_title\} \\
Abstract: \{input\_paper\_text\} \\
\textbf{Output Specification} \\
\begin{verbatim}
{{
    "problem_proximity_score": [0-1 float],
    "methodological_utility_score": [0-1 float],
    "confidence": [0-1 float],
    "reasoning": [Comprehensive explanation with more interpretative flexibility],
    "relevance_score": 0 | 1
}}
\end{verbatim}
Provide \textbf{only} the JSON response. No text outside the JSON object.
\end{tcolorbox}

\begin{tcolorbox}[
    colback=white,
    colframe=pink,
    title=\textbf{\textcolor{black}{MIR-Agent: Proposal Analysis Prompt}},
    boxrule=0.5mm,
    width=\linewidth,
    arc=2mm,
    boxsep=5pt,
    left=6pt, right=6pt, top=6pt, bottom=6pt,
    breakable
]
You will be presented with a Research Proposal.\\
Your task consists of the following steps:\\
1. Analyze the Research Proposal:\\
    - Carefully read each sentence of the Research Proposal.\\
    - Identify and list out core problems Research Proposal aims to solve.\\
    - Very briefly list\\
        1. The main objective and motivation\\
        2. The core problem\\
        3. The sub-problems of the core problem\\
        4. A generic plan of action to solve the core problem\\  
Here is the Research Proposal:
\{query\}
\end{tcolorbox}

\begin{tcolorbox}[
    colback=white,
    colframe=pink,
    title=\textbf{\textcolor{black}{MIR-Agent: Paper Analysis Prompt}},
    boxrule=0.5mm,
    width=\linewidth,
    arc=2mm,
    boxsep=5pt,
    left=6pt, right=6pt, top=6pt, bottom=6pt,
    breakable
]
You are an expert researcher in the field of NLP with rich domain knowledge. \\
Your task is to evaluate the methodological relevance of a \{input\_text\_name\} to a Research Proposal based on the provided criteria. \\
You need to assess whether the methodologies proposed in the\{input\_text\_name\} can contribute towards solving the problem outlined in the \{input\_text\_name\}. \\
Additionally, you will be provided an analysis of the \{input\_text\_name\}.\\

\# Core Assessment Objective \\
Determine if the input paper offers \textbf{substantive methodological utility} that can be: \\
Used/Extended/Adapted for solving the problem stated in the Research Proposal.\\

\#\# Methodological Relevance Criteria

- \textbf{Problem Proximity}\\
  - Addressing Similar research problems.\\
  - Potential for adaptable solution approaches.\\
  - Providing potentially transferable techniques.\\
  - Evaluate if the \{input\_text\_name\} can address any of the sub-problems provided in the analysis of the \{Research Proposal\}.\\
  
- \textbf{Methodological Utility}\\
  - Practical application potential of the \{input\_text\_name\}'s methodologies to the proposal's context.\\
  - Complementarity: Do the methods fill gaps or enhance the proposal's approach?\\
  - Generalizability: Can the methods be extended or adapted to the problem setting of the proposal?\\
  - Evaluate whether the methods offer fresh perspectives or synergize with the proposal's needs.\\
  - Evaluate if the \{input\_text\_name\} can address any part of the plan of action provided in the analysis of the \{Research Proposal\}.\\

Here is the Research Proposal: \\
\{proposal\} \\
Here is the analysis of the Research Proposal: \\
\{proposal\_analysis\} \\
Here is the \{input\_text\_name\}: \\
\{paper\_or\_abstract\}
\end{tcolorbox}

\begin{tcolorbox}[
    colback=white,
    colframe=pink,
    title=\textbf{\textcolor{black}{MIR-Agent: Relevance Judgement Prompt}},
    boxrule=0.5mm,
    width=\linewidth,
    arc=2mm,
    boxsep=5pt,
    left=6pt, right=6pt, top=6pt, bottom=6pt,
    breakable
]
You are an expert researcher in the field of NLP with rich domain knowledge.\\

You will be presented with a Research Proposal, a \{input\_text\_name\}, and an analysis of the \{input\_text\_name\}.\\
Your task is to provide a single one-word judgment whether the \{input\_text\_name\} is methodologically relevant to the Research Proposal based on the analysis and the following criteria.\\

Classify as \textbf{Relevant (1)} if: \\
- The problems of both the proposal and input paper are conceptually related. \\
- The input paper offers baselines to build upon or foundational models/architectures/evaluation metrics that could be adapted. \\
- There is potential for a direct knowledge transfer across domains. \\

Classify as \textbf{Non-Relevant (0)} if: \\
- Orthogonal research domains and no discernible problem relationship. \\
- The input paper offers no clear methodological utility. \\
- Prohibitively complex methodology transfer.\\

Important: Respond using only one of the following two words: \{\{Yes\}\} or \{\{No\}\}.\\

Here is the Research Proposal: \{proposal\} \\
Here is the analysis of the Research Proposal: \{proposal\_analysis\} \\
Here is the \{input\_text\_name\}: \{paper\_or\_abstract\} \\
Here is the analysis of the \{input\_text\_name\}: \{paper\_analysis\} \\

Output: Provide a simple \{\{Yes\}\} or \{\{No\}\} as your answer. \\

Do not include any text outside the JSON object in your response. Provide a simple \{\{Yes\}\} or \{\{No\}\} as your answer.
\end{tcolorbox}
\section*{C Qualitative Analysis}

Following Figure \ref{fig:1}, we use \textcolor{myPastelBlue}{\textbf{blue}} to denote seed paper, \textcolor{myPastelGreen}{\textbf{green}} to denote the ground truth methodologically relevant paper, and \textcolor{myPastelRed}{\textbf{red}} to denote non-methodologically relevant papers.

\subsection*{C.1 Effect of MAG Guided Sampling} \label{C.1}
We observe an improvement in the rankings of methodologically relevant ground truth papers when we use a model fine-tuned with MAG Guided Negative Sampling, as compared to Random Sampling. Specifically, we find that \textbf{15} samples show better rankings. Furthermore, we observe an average rank improvement of \textbf{5.46}.

Our primary motivation for sampling from hard negatives was to push away papers that are semantically related but not methodologically applicable. To validate this, we compare the semantic similarity of the top-k retrieved papers' abstracts (where k=20) with the proposed paper. The average semantic similarity of the top-k papers to the proposal decreased by \textbf{3\%} when using MAG Guided Sampling compared to Random Sampling. To highlight this, we provide the following example:

\begin{tcolorbox}[
    colback=white,
    colframe=myPastelBlue,
    coltext=black,
    title=\textbf{\textcolor{black}{Do you know that Florence is packed with visitors?
Evaluating state-of-the-art models of speaker commitment}},
    boxrule=0.5mm,
    width=\linewidth,
    arc=2mm,
    boxsep=5pt,
    left=6pt, right=6pt, top=6pt, bottom=6pt,
    breakable
]
\textbf{Proposal: }The research aims to understand the challenges in inferring speaker commitment, specifically how current models struggle with different linguistic structures and fail to generalize well to diverse natural language constructions. The motivation is to improve existing speaker commitment models by understanding the linguistic factors contributing to their errors, particularly in handling various linguistic constructions, and ultimately to enhance information extraction and question-answering capabilities in natural language processing.
\end{tcolorbox}

\begin{tcolorbox}[
    colback=white,
    colframe=myPastelGreen,
    title=\textbf{\textcolor{black}{Neural Models of Factuality}},
    boxrule=0.5mm,
    width=\linewidth,
    arc=2mm,
    boxsep=5pt,
    left=6pt, right=6pt, top=6pt, bottom=6pt,
    breakable
]
\textbf{Original Rank: }122 \\
\textbf{Improved Rank: }97 \\\\
\textbf{Abstract: }We present two neural models for event factuality prediction, which yield significant performance gains over previous models on three event factuality datasets: FactBank, UW, and MEANTIME. We also present a substantial expansion of the It Happened portion of the Universal Decompositional Semantics dataset, yielding the largest event factuality dataset to date. We report model results on this extended factuality dataset as well.
\end{tcolorbox}

\begin{tcolorbox}[
    colback=white,
    colframe=myPastelRed,
    title=\textbf{\textcolor{black}{GLUE: A Multi-Task Benchmark and Analysis Platform for Natural Language Understanding}},
    boxrule=0.5mm,
    width=\linewidth,
    arc=2mm,
    boxsep=5pt,
    left=6pt, right=6pt, top=6pt, bottom=6pt,
    breakable
]
\textbf{Original Rank: }5 \\
\textbf{Improved Rank: }10 \\\\
\textbf{Abstract: }Human ability to understand language is general, flexible, and robust. In contrast, most NLU models above the word level are designed for a specific task and struggle with out-of-domain data. If we aspire to develop models with understanding beyond the detection of superficial correspondences between inputs and outputs, then it is critical to develop a unified model that can execute a range of \textcolor{myPastelRed}{\textbf{linguistic}} tasks across different domains. To facilitate research in this direction, we present the General Language Understanding Evaluation (GLUE, gluebenchmark.com): a benchmark of nine diverse NLU tasks, an auxiliary dataset for probing models for understanding specific linguistic phenomena, and an online platform for evaluating and comparing models. For some benchmark tasks, training data is plentiful, but for others it is limited or does not match the genre of the test set. GLUE thus favors models that can represent \textcolor{myPastelRed}{\textbf{linguistic}} knowledge in a way that facilitates sample-efficient learning and effective knowledge-transfer across tasks. While none of the datasets in GLUE were created from scratch for the benchmark, four of them feature privately-held test data, which is used to ensure that the benchmark is used fairly. We evaluate baselines that use ELMo (Peters et al., 2018), a powerful transfer learning technique, as well as state-of-the-art sentence representation models. The best models still achieve fairly low absolute scores. Analysis with our diagnostic dataset yields similarly weak performance over all phenomena tested, with some exceptions.
\end{tcolorbox}

\begin{tcolorbox}[
    colback=white,    
    colframe=black!40!white,  
    title=\textbf{Analysis},
    boxrule=0.5mm,
    width=\linewidth,
    arc=2mm,
    boxsep=5pt,
    left=6pt, right=6pt, top=6pt, bottom=6pt,
    breakable
]
The original paper benchmarks state-of-the-art models on speaker commitment, with \emph{Neural Models of Factuality'} being one of the two key models evaluated. Although the paper’s main objective is to evaluate existing models and their challenges, with the ground truth paper providing the foundational models, it surprisingly ranks very low. One nuance that is difficult to capture here is the use of the term "speaker commitment" which is also referred to as `event factuality', which is used in the ground truth correct paper. Our retriever improves the rank of this paper by \textbf{25}, demonstrating how our model can prioritize methodologically relevant papers, even in the absence of significant semantic overlap.

Conversely, our model also effectively lowers the rankings of papers that share keywords but contribute little to the paper’s topic, highlighting its ability to distinguish between semantically similar but methodologically irrelevant papers.
\end{tcolorbox}

\subsection*{C.2 Effect of Data Augmentation} \label{C.2}
We note \textbf{32} instances where the rankings of ground truth methodologically relevant ranking improved with an average improvement of 17.48, when we employed retrieved through Stella 400M finetuned w/ MIR-MultiCite over Stella 400M fine-tuned w/ MIR-MultiCite-Aug.
\vspace{1em}
\begin{tcolorbox}[
    colback=white,
    colframe=myPastelBlue,
    coltext=black,
    title=\textbf{\textcolor{black}{Do all Roads Lead to Rome?
Understanding the Role of Initialization in Iterative Back-Translation}},
    boxrule=0.5mm,
    width=\linewidth,
    arc=2mm,
    boxsep=5pt,
    left=6pt, right=6pt, top=6pt, bottom=6pt,
    breakable
]
\textbf{Proposal: }The research aims to understand the impact of different initialization methods on the performance of \textcolor{blue}{\textbf{iterative back-translarankstion}} in Neural Machine Translation. Specifically, it investigates whether the final performance is heavily dependent on the initialization quality or if the \textcolor{blue}{\textbf{iterative}} process converges to a similar solution regardless of the initial conditions. The research is motivated by the prevalent use of pre-training or separate machine translation systems to initialize \textcolor{blue}{\textbf{iterative back-translation}} in Neural Machine Translation.
\end{tcolorbox}

\begin{tcolorbox}[
    colback=white,
    colframe=myPastelGreen,
    title=\textbf{\textcolor{black}{An Effective Approach to Unsupervised Machine Translation
}},
    boxrule=0.5mm,
    width=\linewidth,
    arc=2mm,
    boxsep=5pt,
    left=6pt, right=6pt, top=6pt, bottom=6pt,
    breakable
]
\textbf{Original Rank: }7 \\
\textbf{Improved Rank: }1 \\\\
\textbf{Abstract: }While machine translation has traditionally relied on large amounts of parallel corpora, a recent research line has managed to train both Neural Machine Translation (NMT) and Statistical Machine Translation (SMT) systems using monolingual corpora only. In this paper, we identify and address several deficiencies of existing unsupervised SMT approaches by exploiting subword information, developing a theoretically well-founded unsupervised tuning method, and incorporating a joint refinement procedure. Moreover, we use our improved SMT system to initialize a dual NMT model, which is further fine-tuned through \textcolor{darkgreen}{\textbf{on-the-fly back-translation}}. Together, we obtain large improvements over the previous state-of-the-art in unsupervised machine translation. For instance, we get 22.5 BLEU points in English-to-German WMT 2014, 5.5 points more than the previous best-unsupervised system, and 0.5 points more than the (supervised) shared task winner back in 2014.
\end{tcolorbox}

\begin{tcolorbox}[
    colback=white,
    colframe=myPastelRed,
    title=\textbf{\textcolor{black}{Mask-Predict: Parallel Decoding of Conditional Masked Language Models}},
    boxrule=0.5mm,
    width=\linewidth,
    arc=2mm,
    boxsep=5pt,
    left=6pt, right=6pt, top=6pt, bottom=6pt,
    breakable
]
\textbf{Original Rank: }2 \\
\textbf{Improved Rank: }5 \\\\
\textbf{Abstract: }Most \textcolor{myPastelRed}{\textbf{machine translation}} systems generate text autoregressively from left to right. We, instead, use a masked language modeling objective to train a model to predict any subset of the target words, conditioned on both the input text and a partially masked target translation. This approach allows for efficient \textcolor{myPastelRed}{\textbf{iterative}} decoding, where we first predict all of the target words non-autoregressively, and then repeatedly mask out and regenerate the subset of words that the model is least confident about. By applying this strategy for a constant number of \textcolor{myPastelRed}{\textbf{iterations}}, our model improves state-of-the-art performance levels for non-autoregressive and parallel decoding translation models by over 4 BLEU on average. It is also able to reach within about 1 BLEU point of a typical left-to-right transformer model, while decoding significantly faster.
\end{tcolorbox}

\begin{tcolorbox}[
    colback=white,    
    colframe=black!40!white,  
    title=\textbf{Analysis},
    boxrule=0.5mm,
    width=\linewidth,
    arc=2mm,
    boxsep=5pt,
    left=6pt, right=6pt, top=6pt, bottom=6pt,
    breakable
]
The proposal suggests the use of Neural Machine Translation (NMT) with iterative back-translation, the primary objective is to investigate whether initialization methods significantly impact performance. This approach requires existing models capable of performing MT using iterative back-translation within a dual NMT framework. A key aspect from the proposal, mentioned thrice is the keyword `iterative back-translation', but the correct methodology relevant paper mentions a slight paraphrase using `on-the-fly back-translation', while the wrongly retrieved paper mentions `iterative', it holds no significance and is completely disconnected from the problem mentioned in the proposal and instead aims to create a MT system which is not autoregressive by pre-training on a MLM objective, attaining faster decoding. We hypothesize that since abstract mentions the keyword `iterative' thrice, biasing our retriever trained on original training data to prioritise this work. Where as post-training with our augmented data, we highlight the improvement in rank of the ground truth and decrease in the rank of the incorrect sample.
\\
\textbf{Note:} Original Train set has 1 mention of `iterative back-translation' where as augmented has 154 mentions of the same, highlighting the positive effects of having consistent domain representation in the training set to be essential for downstream MIR performance.
\end{tcolorbox}

\subsection*{C.3 Corrections by Zero Shot LLM Re-ranking} \label{C.3}
We qualitatively analyze the samples where LLM was able to improve ranks of ground truth correct labels by identifying retrievals above ground truth as non-methodologically relevant.
\vspace{1em}
\begin{tcolorbox}[
    colback=white,
    colframe=myPastelBlue,
    coltext=black,
    title=\textbf{\textcolor{black}{Modeling Long-Range Context for
Concurrent Dialogue Acts Recognition}},
    boxrule=0.5mm,
    width=\linewidth,
    arc=2mm,
    boxsep=5pt,
    left=6pt, right=6pt, top=6pt, bottom=6pt,
    breakable
]
\textbf{Proposal: }The research problem is the challenge of recognizing and analyzing `Concurrent Dialogue Acts' (CDA) in dialogues. CDA occurs when a single utterance within a dialogue expresses multiple dialogue acts, leading to complex dependencies across dialogue turns. Existing approaches to dialogue act recognition struggle with this challenge, either assuming only one dialogue act per utterance or failing to account for the sequential nature of dialogues. The research is motivated by the limitations of existing methods for dialogue act recognition, which fail to handle the complexity of utterances containing multiple dialogue acts, especially within long-range dialogue contexts.
\end{tcolorbox}

\begin{tcolorbox}[
    colback=white,
    colframe=myPastelRed,
    title=\textbf{\textcolor{black}{Robust dialog state tracking using delexicalised recurrent neural networks and unsupervised adaptation}},
    boxrule=0.5mm,
    width=\linewidth,
    arc=2mm,
    boxsep=5pt,
    left=6pt, right=6pt, top=6pt, bottom=6pt,
    breakable
]
\textbf{Retriever Rank: }1 \\
\textbf{LLM Rank: }1 \\
\textbf{Abstract: }Tracking the user's intention throughout the course of a dialog, called dialog state tracking, is an important component of any dialog system. Most existing spoken dialog systems are designed to work in a static, well-defined domain, and are not well suited to tasks in which the domain may change or be extended over time. This paper shows how recurrent neural networks can be effectively applied to tracking in an extended domain with new slots and values not present in training data. The method is evaluated in the third Dialog State Tracking Challenge, where it significantly outperforms other approaches in the task of tracking the user's goal. A method for online unsupervised adaptation to new domains is also presented. Unsupervised adaptation is shown to be helpful in improving word-based recurrent neural networks, which work directly from the speech recognition results. Word-based dialog state tracking is attractive as it does not require engineering a spoken language understanding system for use in the new domain and it avoids the general-purpose purpose intermediate semantic representation.
\end{tcolorbox}

\begin{tcolorbox}[
    colback=white,
    colframe=pink,
    title=\textbf{\textcolor{black}{LLM Reasoning}},
    boxrule=0.5mm,
    width=\linewidth,
    arc=2mm,
    boxsep=5pt,
    left=6pt, right=6pt, top=6pt, bottom=6pt,
    breakable
]
"problem\_proximity\_score": 0.4,\\
"methodological\_utility\_score": 0.3,\\
"confidence": 0.6,\\
"reasoning": "The input paper tackles dialog state tracking, which is related to dialogue act recognition but focuses on understanding user intention within a dialogue's evolving context. While not directly addressing Concurrent Dialogue Acts, the paper's focus on handling new slots and values in dynamic domains holds some relevance. The RNN-based approach could potentially be adapted to model the sequential dependencies present in CDAs, but this would require significant modification. The paper's strength lies in handling domain changes, which is not the primary concern of the CDA problem.  The unsupervised adaptation aspect might be useful for handling variability in CDA expressions, but its applicability remains uncertain without further investigation.",\\
"relevance\_score": 1
\end{tcolorbox}

\begin{tcolorbox}[
    colback=white,
    colframe=myPastelRed,
    title=\textbf{\textcolor{black}{The Ubuntu Dialogue Corpus: A Large Dataset for Research in Unstructured Multi-Turn Dialogue Systems}},
    boxrule=0.5mm,
    width=\linewidth,
    arc=2mm,
    boxsep=5pt,
    left=6pt, right=6pt, top=6pt, bottom=6pt,
    breakable
]
\textbf{Retriever Rank: }2 \\
\textbf{LLM Rank: }3 \\
\textbf{Abstract: }This paper introduces the Ubuntu Dialogue Corpus, a dataset containing almost 1 million multi-turn dialogues, with a total of over 7 million utterances and 100 million words. This provides a unique resource for research into building dialogue managers based on neural language models that can make use of large amounts of unlabeled data. The dataset has both the multi-turn property of conversations in the Dialog State Tracking Challenge datasets, and the unstructured nature of interactions from microblog services such as Twitter. We also describe two neural learning architectures suitable for analyzing this dataset, and provide benchmark performance on the task of selecting the best next response.
\end{tcolorbox}

\begin{tcolorbox}[
    colback=white,
    colframe=pink,
    title=\textbf{\textcolor{black}{LLM Reasoning}},
    boxrule=0.5mm,
    width=\linewidth,
    arc=2mm,
    boxsep=5pt,
    left=6pt, right=6pt, top=6pt, bottom=6pt,
    breakable
]
"problem\_proximity\_score": 0.2,\\
"methodological\_utility\_score": 0.1,\\
"confidence": 0.6,\\
"reasoning": "While both the proposal and input paper operate within the domain of dialogue analysis, their core problems differ significantly. The proposal focuses on recognizing 'Concurrent Dialogue Acts' within utterances, highlighting the challenge of multiple acts within a single turn. Conversely, the input paper introduces a dataset for 'unstructured multi-turn dialogue systems' focusing on broader conversational flow and response selection, \textcolor{myPastelRed}{\textbf{not fine-grained dialogue act analysis}}. The input paper's methodological utility is limited for the proposal. While the large-scale dataset could potentially be explored for instances of CDAs, its \textcolor{myPastelRed}{\textbf{annotation scheme is not designed for this task}}. The neural architectures proposed are for \textcolor{myPastelRed}{\textbf{response selection, not CDA recognition.}} Thus, direct application or adaptation is unlikely. There's a slight chance the dataset could be useful for extracting features relevant to CDA, and the neural models might provide a starting point for architectures handling multi-act utterances. However, these are weak connections, and significant adaptation would be needed.",\\
"relevance\_score": 0
\end{tcolorbox}

\begin{tcolorbox}[
    colback=white,
    colframe=myPastelRed,
    title=\textbf{\textcolor{black}{Learning to Disentangle Interleaved Conversational Threads with a Siamese Hierarchical Network and Similarity Ranking}},
    boxrule=0.5mm,
    width=\linewidth,
    arc=2mm,
    boxsep=5pt,
    left=6pt, right=6pt, top=6pt, bottom=6pt,
    breakable
]
\textbf{Retriever Rank: }3 \\
\textbf{LLM Rank: }4 \\
\textbf{Abstract: }An enormous amount of conversation occurs online every day, such as on chat platforms where multiple conversations may take place concurrently. Interleaved conversations lead to difficulties in not only following discussions but also retrieving relevant information from simultaneous messages. Conversation disentanglement aims to separate intermingled messages into detached conversations. In this paper, we propose to leverage representation learning for conversation disentanglement. A Siamese hierarchical convolutional neural network (SHCNN), which integrates local and more global representations of a message, is first presented to estimate the conversation-level similarity between closely posted messages. With the estimated similarity scores, our algorithm for conversation identification by similarity ranking (CISIR) then derives conversations based on high-confidence message pairs and pairwise redundancy. Experiments were conducted with four publicly available datasets of conversations from Reddit and IRC channels. The experimental results show that our approach significantly outperforms comparative baselines in both pairwise similarity estimation and conversation disentanglement.
\end{tcolorbox}

\begin{tcolorbox}[
    colback=white,
    colframe=pink,
    title=\textbf{\textcolor{black}{LLM Reasoning}},
    boxrule=0.5mm,
    width=\linewidth,
    arc=2mm,
    boxsep=5pt,
    left=6pt, right=6pt, top=6pt, bottom=6pt,
    breakable
]
"problem\_proximity\_score": 0.4,\\
"methodological\_utility\_score": 0.2,\\
"confidence": 0.3,\\
"reasoning": "While both the proposal and the input paper deal with analyzing dialogues, their core problems are not closely aligned. The proposal focuses on recognizing multiple dialogue acts within single utterances (CDA), while the paper addresses disentangling interleaved conversations. However, there might be some limited methodological utility. The SHCNN architecture used in the paper for learning message similarity could potentially be adapted to learn representations of utterances that capture the presence of multiple dialogue acts. Similarly, the concept of 'similarity ranking' used for disentanglement might offer a fresh perspective on identifying dependencies between dialogue acts within a single utterance. Despite these potential connections, the transferability of the methodologies is not straightforward. The paper's focus on separating entire conversations is quite \textcolor{myPastelRed}{\textbf{different from analyzing the nuances of single utterances with multiple dialogue acts.}} The hierarchical nature of the SHCNN, designed for longer text segments, might be unnecessarily complex for shorter utterances. Additionally, the 'similarity ranking' in the paper relies on the temporal proximity of messages within a conversation, which doesn't directly translate to the CDA problem. Overall, the input paper presents a somewhat distant problem domain with limited methodological overlap. The techniques used, while potentially offering some conceptual inspiration, would require significant adaptation and might not be the most effective approach for tackling the CDA challenge.",\\
"relevance\_score": 0
\end{tcolorbox}

\begin{tcolorbox}[
    colback=white,
    colframe=myPastelGreen,
    title=\textbf{\textcolor{black}{User Intent Prediction in Information-seeking Conversations}},
    boxrule=0.5mm,
    width=\linewidth,
    arc=2mm,
    boxsep=5pt,
    left=6pt, right=6pt, top=6pt, bottom=6pt,
    breakable
]
\textbf{Original Rank: }4 \\
\textbf{Improved Rank: }2 \\\\
\textbf{Abstract: }Conversational assistants are being progressively adopted by the general population. However, they are not capable of handling complicated information-seeking tasks that involve multiple turns of information exchange. Due to the limited communication bandwidth in conversational search, it is important for conversational assistants to accurately detect and predict user intent in information-seeking conversations. In this paper, we investigate two aspects of user intent prediction in an information-seeking setting. First, we extract features based on the content, structural, and sentiment characteristics of a given utterance, and use classic machine learning methods to perform user intent prediction. We then conduct an in-depth feature importance analysis to identify key features in this prediction task. We find that structural features contribute most to the prediction performance. Given this finding, we construct neural classifiers to incorporate context information and achieve better performance without feature engineering. Our findings can provide insights into the important factors and effective methods of user intent prediction in information-seeking conversations.
\end{tcolorbox}

\begin{tcolorbox}[
    colback=white,
    colframe=pink,
    title=\textbf{\textcolor{black}{LLM Reasoning}},
    boxrule=0.5mm,
    width=\linewidth,
    arc=2mm,
    boxsep=5pt,
    left=6pt, right=6pt, top=6pt, bottom=6pt,
    breakable
]
"problem\_proximity\_score": 0.7,\\
"methodological\_utility\_score": 0.6,\\
"confidence": 0.8,\\
"reasoning": "While the input paper focuses on user intent prediction in information-seeking conversations, which is not directly equivalent to recognizing Concurrent Dialogue Acts (CDA), there are significant overlaps and potential for adaptation. \\Problem Proximity: Both the proposal and paper address challenges in understanding complex dialogue flow. \textcolor{darkgreen}{\textbf{Recognizing user intent within an information-seeking dialogue can be seen as a specific case of identifying dialogue acts.}} Furthermore, the paper acknowledges the importance of context in understanding user intent, aligning with the proposal's emphasis on sequential dependencies in CDA recognition. \\Methodological Utility: The input paper's focus on feature extraction (content, structural, sentiment) from utterances provides a valuable starting point for CDA analysis. Although the specific features might need adaptation for CDA recognition, the underlying principle of leveraging diverse utterance characteristics remains relevant. Additionally, the paper's exploration of neural classifiers to incorporate context information directly addresses the proposal's concern about capturing long-range dependencies in dialogues. \\Limitations: The paper's focus on user intent prediction in information-seeking dialogues limits the direct transferability of its methodologies. CDA recognition might involve a broader range of dialogue acts and require different contextual considerations. However, the input paper provides a solid foundation for building upon and adapting its methods for the more general problem of CDA recognition.",\\
"relevance\_score": 1
\end{tcolorbox}

\begin{tcolorbox}[
    colback=white,    
    colframe=black!40!white,  
    title=\textbf{Analysis},
    boxrule=0.5mm,
    width=\linewidth,
    arc=2mm,
    boxsep=5pt,
    left=6pt, right=6pt, top=6pt, bottom=6pt,
    breakable
]
This example demonstrates how LLMs, as strong basic common-sense reasoners, can effectively filter out papers that address orthogonal problems or domains—something that purely retrieval-based methods can fail at.\\

Specifically, we highlight instances where the LLM is able to identify that (1) the provided paper focuses on broader conversational flow and the provided unlabelled dataset is not suited for the fine-grained dialogue act analysis and (2) the problem of separating entire conversations is quite different from the need of the proposal, and not directly adaptable or extendable.
\end{tcolorbox}

\subsection*{C.4 Effect of providing Few Shots} \label{C.4}
We qualitatively analyze the impact of providing a few-shot learning approach with citation contexts, which offers the LLM a more flexible demonstration of what is methodologically relevant. In the following example, we show how, with effective few-shot examples, the LLM's reasoning improves, leading to better re-ranking. Specifically, the model was able to filter out several non-relevant papers, correctly positioning the ground-truth relevant paper higher in the ranking. In contrast, without few-shot learning, all papers were mistakenly classified as relevant, failing to improve the ranking of the ground-truth relevant paper.
\vspace{1em}
\begin{tcolorbox}[
    colback=white,
    colframe=myPastelBlue,
    coltext=black,
    title=\textbf{\textcolor{black}{Analyzing ASR Pretraining for Low-Resource Speech-to-Text Translation}},
    boxrule=0.5mm,
    width=\linewidth,
    arc=2mm,
    boxsep=5pt,
    left=6pt, right=6pt, top=6pt, bottom=6pt,
    breakable
]
\textbf{Proposal: }The research focuses on improving automatic speech-to-text translation (AST) for low-resource source languages. Specifically, it investigates the impact of factors like language relatedness and dataset size on pre-training's effectiveness and explores the potential of combining pre-training with data augmentation. The research aims to shed light on the effectiveness of pre-training in AST for low-resource languages and understand the influence of different factors on its performance. It also seeks to explore the potential benefits of combining pre-training with data augmentation.
\end{tcolorbox}

\begin{tcolorbox}[
    colback=white,
    colframe=myPastelRed,
    title=\textbf{\textcolor{black}{Unsupervised Word Segmentation from Speech with Attention}},
    boxrule=0.5mm,
    width=\linewidth,
    arc=2mm,
    boxsep=5pt,
    left=6pt, right=6pt, top=6pt, bottom=6pt,
    breakable
]
\textbf{Original Rank: }1 \\
\textbf{LLM Rank: }5 \\\\
\textbf{Abstract: }We present a first attempt to perform attentional word segmentation directly from the speech signal, with the final goal to automatically identify lexical units in a low-resource, unwritten language (UL). Our methodology assumes a pairing between recordings in the UL with translations in a well-resourced language. It uses Acoustic Unit Discovery (AUD) to convert speech into a sequence of pseudo-phones that is segmented using neural soft-alignments produced by a neural machine translation model. Evaluation uses an actual Bantu UL, Mboshi; comparisons to monolingual and bilingual baselines illustrate the potential of attentional word segmentation for language documentation.
\end{tcolorbox}

\begin{tcolorbox}[
    colback=white,
    colframe=pink,
    title=\textbf{\textcolor{black}{LLM Reasoning}},
    boxrule=0.5mm,
    width=\linewidth,
    arc=2mm,
    boxsep=5pt,
    left=6pt, right=6pt, top=6pt, bottom=6pt,
    breakable
]
"problem\_proximity\_score": 0.6,\\
"methodological\_utility\_score": 0.4,\\
"confidence": 0.7,\\
"reasoning": "The input paper tackles word segmentation in low-resource unwritten languages, which is relevant to the proposal's focus on low-resource AST. However, the paper focuses on \textcolor{myPastelRed}{\textbf{identifying lexical units using acoustic features and translations, while the proposal aims to improve speech-to-text translation itself.}} The methodologies are not directly transferable, as the proposal likely requires techniques for mapping speech to target language text, not just word segmentation. While the attention-based segmentation approach might offer insights for improving alignment in AST, the paper's focus on acoustic features limits its direct utility for the proposal's problem.  The input paper's focus on low-resource languages and use of neural models makes it somewhat relevant, but the methodological gap between word segmentation and translation limits its direct applicability.",\\
"relevance\_score": 0
\end{tcolorbox}

\begin{tcolorbox}[
    colback=white,
    colframe=myPastelRed,
    title=\textbf{\textcolor{black}{Multilingual Denoising Pre-training for Neural Machine Translation}},
    boxrule=0.5mm,
    width=\linewidth,
    arc=2mm,
    boxsep=5pt,
    left=6pt, right=6pt, top=6pt, bottom=6pt,
    breakable
]
\textbf{Original Rank: }2 \\
\textbf{LLM Rank: }1 \\\\
\textbf{Abstract: }This paper demonstrates that multilingual denoising pre-training produces significant performance gains across a wide variety of machine translation (MT) tasks. We present mBART—a sequence-to-sequence denoising auto-encoder pre-trained on large-scale monolingual corpora in many languages using the BART objective (Lewis et al., 2019). mBART is the first method for pre-training a complete sequence-to-sequence model by denoising full texts in multiple languages, whereas previous approaches have focused only on the encoder, decoder, or reconstructing parts of the text. Pre-training a complete model allows it to be directly fine-tuned for supervised (both sentence-level and document-level) and unsupervised machine translation, with no task- specific modifications. We demonstrate that adding mBART initialization produces performance gains in all but the highest-resource settings, including up to 12 BLEU points for low resource MT and over 5 BLEU points for many document-level and unsupervised models. We also show that it enables transfer to language pairs with no bi-text or that were not in the pre-training corpus, and present extensive analysis of which factors contribute the most to effective pre-training.
\end{tcolorbox}

\begin{tcolorbox}[
    colback=white,
    colframe=pink,
    title=\textbf{\textcolor{black}{LLM Reasoning}},
    boxrule=0.5mm,
    width=\linewidth,
    arc=2mm,
    boxsep=5pt,
    left=6pt, right=6pt, top=6pt, bottom=6pt,
    breakable
]
"problem\_proximity\_score": 0.7,\\
"methodological\_utility\_score": 0.9,\\
"confidence": 0.8,\\
"reasoning": "The input paper, while focusing on multilingual denoising pre-training for general MT, holds substantial methodological relevance to the proposal's focus on low-resource AST.  Here's why:\\Problem Proximity: Both target Neural Machine Translation, with the proposal specifically honing in on the challenges within low-resource Automatic Speech-To-Text translation. This makes the problems closely aligned, even if not perfectly identical. \\Methodological Utility:\\    - mBART Adaptability: The paper's mBART model, pre-trained on large-scale monolingual corpora, offers a strong foundation for the proposal.  It can be directly fine-tuned for AST, potentially even with the added noise inherent in speech data, to investigate its efficacy in the low-resource setting.\\    - Data Augmentation Synergy: The proposal's interest in data augmentation neatly complements the paper's findings.  The success of mBART's pre-training suggests that augmenting low-resource AST data, perhaps with techniques like back-translation informed by mBART, could yield significant improvements.\\    - Analysis Framework: The paper's analysis of factors influencing pre-training effectiveness provides a valuable starting point for the proposal. This framework can be adapted to explore how language relatedness and dataset size specifically impact pre-training for low-resource AST.\\Relevance:  The input paper doesn't directly address AST, but its focus on multilingual denoising pre-training and analysis of pre-training factors offers substantial methodological utility for the proposal.  The techniques are adaptable, and the findings directly inform the proposal's research questions. Therefore, it is highly relevant.",\\
"relevance\_score": 1
\end{tcolorbox}
\begin{tcolorbox}[
    colback=white,
    colframe=myPastelRed,
    title=\textbf{\textcolor{black}{A Very Low Resource Language Speech Corpus for Computational Language Documentation Experiments}},
    boxrule=0.5mm,
    width=\linewidth,
    arc=2mm,
    boxsep=5pt,
    left=6pt, right=6pt, top=6pt, bottom=6pt,
    breakable
]
\textbf{Original Rank: }3 \\
\textbf{LLM Rank: }6 \\\\
\textbf{Abstract: }Most speech and language technologies are trained with massive amounts of speech and text information. However, most of the world's languages do not have such resources or stable orthography. Systems constructed under these almost zero resource conditions are not only promising for speech technology but also for computational language documentation. The goal of computational language documentation is to help field linguists to (semi-)automatically analyze and annotate audio recordings of endangered and unwritten languages. Example tasks are automatic phoneme discovery or lexicon discovery from the speech signal. This paper presents a speech corpus collected during a realistic language documentation process. It is made up of 5k speech utterances in Mboshi (Bantu C25) aligned to French text translations. Speech transcriptions are also made available: they correspond to a non-standard graphemic form close to the language phonology. We present how the data was collected, cleaned and processed and we illustrate its use through a zero-resource task: spoken term discovery. The dataset is made available to the community for reproducible computational language documentation experiments and their evaluation.
\end{tcolorbox}

\begin{tcolorbox}[
    colback=white,
    colframe=pink,
    title=\textbf{\textcolor{black}{LLM Reasoning}},
    boxrule=0.5mm,
    width=\linewidth,
    arc=2mm,
    boxsep=5pt,
    left=6pt, right=6pt, top=6pt, bottom=6pt,
    breakable
]
"problem\_proximity\_score": 0.2,\\
"methodological\_utility\_score": 0.1,\\
"confidence": 0.7,\\
"reasoning": "The research proposal and the input paper both address low-resource languages, which signifies some problem proximity. However, the proposal focuses on improving speech-to-text translation (AST) through pre-training and data augmentation, while the paper focuses on creating a speech corpus for computational language documentation tasks like phoneme and lexicon discovery. This difference in focus makes the methodologies presented in the paper less directly applicable to the proposal's problem. While the corpus creation process and the spoken term discovery task might offer some insights, the paper lacks direct methodological contributions to pre-training or data augmentation for AST. Therefore, the paper holds limited methodological utility for the proposal, making it less relevant.",\\
"relevance\_score": 0
\end{tcolorbox}

\begin{tcolorbox}[
    colback=white,
    colframe=myPastelRed,
    title=\textbf{\textcolor{black}{Sequence-to-Sequence Models Can Directly Translate Foreign Speech}},
    boxrule=0.5mm,
    width=\linewidth,
    arc=2mm,
    boxsep=5pt,
    left=6pt, right=6pt, top=6pt, bottom=6pt,
    breakable
]
\textbf{Original Rank: }4 \\
\textbf{LLM Rank: }2 \\\\
\textbf{Abstract: }We present a recurrent encoder-decoder deep neural network architecture that directly translates speech in one language into text in another. The model does not explicitly transcribe the speech into text in the source language, nor does it require supervision from the ground truth source language transcription during training. We apply a slightly modified sequence-to-sequence with attention architecture that has previously been used for speech recognition and show that it can be repurposed for this more complex task, illustrating the power of attention-based models. A single model trained end-to-end obtains state-of-the-art performance on the Fisher Callhome Spanish-English speech translation task, outperforming a cascade of independently trained sequence-to-sequence speech recognition and machine translation models by 1.8 BLEU points on the Fisher test set. In addition, we find that making use of the training data in both languages by multi-task training sequence-to-sequence speech translation and recognition models with a shared encoder network can improve performance by a further 1.4 BLEU points.
\end{tcolorbox}

\begin{tcolorbox}[
    colback=white,
    colframe=pink,
    title=\textbf{\textcolor{black}{LLM Reasoning}},
    boxrule=0.5mm,
    width=\linewidth,
    arc=2mm,
    boxsep=5pt,
    left=6pt, right=6pt, top=6pt, bottom=6pt,
    breakable
]
"problem\_proximity\_score": 0.7,\\
"methodological\_utility\_score": 0.3,\\
"confidence": 0.6,\\
"reasoning": "The input paper tackles direct speech-to-text translation, which is relevant but not identical to the proposal's focus on low-resource AST.  The paper's sequence-to-sequence model and attention mechanism offer a relevant starting point. However, its focus on high-resource scenarios with a shared encoder for multi-task learning might not directly translate to low-resource settings. The paper's findings on direct translation and multi-tasking offer valuable insights but require adaptation for the proposal's specific context.",\\
"relevance\_score": 1
\end{tcolorbox}

\begin{tcolorbox}[
    colback=white,
    colframe=myPastelRed,
    title=\textbf{\textcolor{black}{A Comparative Study on Transformer vs RNN in Speech Applications}},
    boxrule=0.5mm,
    width=\linewidth,
    arc=2mm,
    boxsep=5pt,
    left=6pt, right=6pt, top=6pt, bottom=6pt,
    breakable
]
\textbf{Original Rank: }5 \\
\textbf{LLM Rank: }7 \\\\
\textbf{Abstract: }Sequence-to-sequence models have been widely used in end-to-end speech processing, for example, automatic speech recognition (ASR), speech translation (ST), and text-to-speech (TTS). This paper focuses on an emergent sequence-to-sequence model called Transformer, which achieves state-of-the-art performance in neural machine translation and other natural language processing applications. We undertook intensive studies in which we experimentally compared and analyzed Transformer and conventional recurrent neural networks (RNN) in a total of 15 ASR, one multilingual ASR, one ST, and two TTS benchmarks. Our experiments revealed various training tips and significant performance benefits obtained with Transformer for each task including the surprising superiority of Transformer in 13/15 ASR benchmarks in comparison with RNN. We are preparing to release Kaldi-style reproducible recipes using open source and publicly available datasets for all the ASR, ST, and TTS tasks for the community to succeed our exciting outcomes.
\end{tcolorbox}

\begin{tcolorbox}[
    colback=white,
    colframe=pink,
    title=\textbf{\textcolor{black}{LLM Reasoning}},
    boxrule=0.5mm,
    width=\linewidth,
    arc=2mm,
    boxsep=5pt,
    left=6pt, right=6pt, top=6pt, bottom=6pt,
    breakable
]
"problem\_proximity\_score": 0.3,\\
"methodological\_utility\_score": 0.2,\\
"confidence": 0.6,\\
"reasoning": "The input paper, while focusing on speech applications like the proposal, primarily compares Transformer and RNN architectures for various speech tasks. Although it touches upon speech translation (ST), which is relevant to the proposal's focus on automatic speech-to-text translation (AST), it \textcolor{myPastelRed}{\textbf{doesn't delve into the low-resource aspect that is central to the proposal.}} \\The paper's exploration of Transformer architectures could offer some methodological utility as a starting point. However, the proposal specifically investigates pre-training and data augmentation for low-resource AST, areas not directly addressed in the input paper.  The paper's focus on established benchmarks may not be entirely relevant to low-resource scenarios. \\Overall, the input paper's findings on Transformer architectures might offer a base, but significant adaptation and extension would be needed to address the proposal's specific concerns. Thus, while not entirely irrelevant, the methodological utility is limited.",\\
"relevance\_score": 0
\end{tcolorbox}

\begin{tcolorbox}[
    colback=white,
    colframe=myPastelRed,
    title=\textbf{\textcolor{black}{SpecAugment: A Simple Data Augmentation Method for Automatic Speech Recognition}},
    boxrule=0.5mm,
    width=\linewidth,
    arc=2mm,
    boxsep=5pt,
    left=6pt, right=6pt, top=6pt, bottom=6pt,
    breakable
]
\textbf{Original Rank: }6 \\
\textbf{LLM Rank: }3 \\\\
\textbf{Abstract: }We present SpecAugment, a simple data augmentation method for speech recognition. SpecAugment is applied directly to the feature inputs of a neural network (i.e., filter bank coefficients). The augmentation policy consists of warping the features, masking blocks of frequency channels, and masking blocks of time steps. We apply SpecAugment on Listen, Attend and Spell networks for end-to-end speech recognition tasks. We achieve state-of-the-art performance on the LibriSpeech 960h and Swichboard 300h tasks, outperforming all prior work. On LibriSpeech, we achieve 6.8\% WER on test-other without the use of a language model, and 5.8\% WER with shallow fusion with a language model. This compares to the previous state-of-the-art hybrid system of 7.5\% WER. For Switchboard, we achieve 7.2\%/14.6\% on the Switchboard/CallHome portion of the Hub5'00 test set without the use of a language model, and 6.8\%/14.1\% with shallow fusion, which compares to the previous state-of-the-art hybrid system at 8.3\%/17.3\% WER.
\end{tcolorbox}

\begin{tcolorbox}[
    colback=white,
    colframe=pink,
    title=\textbf{\textcolor{black}{LLM Reasoning}},
    boxrule=0.5mm,
    width=\linewidth,
    arc=2mm,
    boxsep=5pt,
    left=6pt, right=6pt, top=6pt, bottom=6pt,
    breakable
]
"problem\_proximity\_score": 0.6,\\
"methodological\_utility\_score": 0.7,\\
"confidence": 0.8,\\
"reasoning": "While the input paper focuses on automatic speech recognition (ASR) and the proposal targets speech-to-text translation (AST), there is a clear overlap in the data augmentation aspect. SpecAugment's application to speech features could potentially be adapted for AST, especially considering the proposal's interest in data augmentation for low-resource settings. However, direct transferability isn't guaranteed as AST introduces additional complexities with translation. The paper's focus on ASR limits its direct applicability to the translation aspect of the proposal.  The relevance is high in terms of data augmentation techniques for speech data, but the difference in tasks (ASR vs. AST) slightly lowers the overall relevance.",\\
"relevance\_score": 1
\end{tcolorbox}

\begin{tcolorbox}[
    colback=white,
    colframe=myPastelGreen,
    title=\textbf{\textcolor{black}{End-to-End Automatic Speech Translation of Audiobooks}},
    boxrule=0.5mm,
    width=\linewidth,
    arc=2mm,
    boxsep=5pt,
    left=6pt, right=6pt, top=6pt, bottom=6pt,
    breakable
]
\textbf{Original Rank: }7 \\
\textbf{Improved Rank: }4 \\\\
\textbf{Abstract: }We investigate end-to-end speech-to-text translation on a corpus of audiobooks specifically augmented for this task. Previous works investigated the extreme case where source language transcription is not available during learning nor decoding, but we also study a midway case where source language transcription is available at training time only. In this case, a single model is trained to decode source speech into target text in a single pass. Experimental results show that it is possible to train compact and efficient end-to-end speech translation models in this setup. We also distribute the corpus and hope that our speech translation baseline on this corpus will be challenged in the future.
\end{tcolorbox}

\begin{tcolorbox}[
    colback=white,
    colframe=pink,
    title=\textbf{\textcolor{black}{LLM Reasoning}},
    boxrule=0.5mm,
    width=\linewidth,
    arc=2mm,
    boxsep=5pt,
    left=6pt, right=6pt, top=6pt, bottom=6pt,
    breakable
]
"problem\_proximity\_score": 0.6,\\
"methodological\_utility\_score": 0.4,\\
"confidence": 0.7,\\
"reasoning": "The input paper focuses on end-to-end speech-to-text translation using audiobooks, specifically exploring scenarios with and without source language transcription during training. While this differs from the proposal's focus on low-resource AST and pre-training, there are overlapping areas:\\Problem Proximity: Both target automatic speech translation, and the paper's focus on data augmentation (using audiobooks) slightly overlaps with the proposal's interest in data augmentation for low-resource scenarios. However, the paper doesn't directly address low-resource languages or pre-training.\\Methodological Utility: The end-to-end AST model and the investigation of training with/without transcription could offer some adaptable components for the proposal. For instance, the paper's model could potentially be used as a baseline for comparison or adapted for a low-resource setting.  However, the direct applicability of the paper's methods is limited due to the different focuses. \\Overall, the input paper exhibits some problem proximity due to the shared domain of AST and the tangential link with data augmentation. It offers limited but potentially adaptable methodological components. Therefore, it holds some relevance to the proposal, but not a strong one.",\\
"relevance\_score": 1
\end{tcolorbox}

\begin{tcolorbox}[
    colback=white,    
    colframe=purple!40!white,  
    title=\textbf{Few Shots},
    boxrule=0.5mm,
    width=\linewidth,
    arc=2mm,
    boxsep=5pt,
    left=6pt, right=6pt, top=6pt, bottom=6pt,
    breakable
]
Exemplar:\\
Sample Proposal: The research investigates the challenges of limited labeled training data in sequence prediction tasks, particularly for part-of-speech tagging and supertagging. It aims to find effective methods for training models with limited labeled data and even without any labeled examples. The research is motivated by the need to improve the performance of sequence prediction models, especially when dealing with limited labeled training data. The study seeks to find alternative methods to train models effectively, including utilizing unlabeled resources and co-training techniques.\\\\

Paper: Much previous work has investigated weak supervision with HMMs and tag dictionaries for part-of-speech tagging, but there have been no similar investigations for the harder problem of supertagging. Here, I show that weak supervision for supertagging does work, but that it is subject to severe performance degradation when the tag dictionary is highly ambiguous. I show that lexical category complexity and information about how supertags may combine syntactically can be used to initialize the transition distributions of a first-order Hidden Markov Model for weakly supervised learning. This initialization proves more effective than starting with uniform transitions, especially when the tag dictionary is highly ambiguous.\\

Methodology Citations:\\
1. 
Citation: We draw the initial sample of CCG tag sequences corresponding to the observation sequence, using probabilities based on grammar informed initialization (Baldridge, 2008).\\
Relevance Score: 1\\\\
2. 
Citation: In this experiment, we use the training and test sets used by Baldridge (2008) from CCGbank.\\
Relevance Score: 1\\\\
Non-Methodology Citation:\\
Citation: Previous work has used Bayesian HMMs to learn taggers for both POS tagging and supertagging (Baldridge, 2008) separately.\\
Relevance Score: 0

\end{tcolorbox}

\begin{tcolorbox}[
    colback=white,    
    colframe=black!40!white,  
    title=\textbf{Analysis},
    boxrule=0.5mm,
    width=\linewidth,
    arc=2mm,
    boxsep=5pt,
    left=6pt, right=6pt, top=6pt, bottom=6pt,
    breakable
]
In comparison to zero-shot for the same proposal, where the LLM classified all papers as methodologically relevant, but by providing grounding, the LLM becomes stricter which allows the ranking of the ground truth paper to improve.\\\\
Specifically, the exemplar provided illustrates that papers that provide background context for the problem in the proposal, are not specifically aligned to contributing towards building new methodologies. This enables the LLM to filter out samples where the paper provides background for the problem and compares RNNs and Transformers for the task of Speech Translation, it doesn't dive into the central concern of building these systems in low-resource settings. 
\end{tcolorbox}

\subsection*{C.5 Improvement of MIR-Agent over Pointwise Re-ranking} \label{C.5}
To understand how our MIR-Agent pipeline goes through intermediate analysis to arrive at the best judgment improving performance even further, we show the analysis provided by both Pointwise and MIR-Agent pipeline and attempt to interpret the benefits of reasoning-intensive retrieval
\vspace{1em}
\begin{tcolorbox}[
    colback=white,
    colframe=myPastelBlue,
    coltext=black,
    title=\textbf{\textcolor{black}{Abstractive Document Summarization without Parallel Data}},
    boxrule=0.5mm,
    width=\linewidth,
    arc=2mm,
    boxsep=5pt,
    left=6pt, right=6pt, top=6pt, bottom=6pt,
    breakable
]
\textbf{Proposal: }The research focuses on the challenge of abstractive summarization when parallel data, consisting of paired articles and summaries, is scarce and expensive to acquire. The research is motivated by the need to address the limitations of traditional abstractive summarization methods that rely heavily on parallel data, which is often scarce and costly. The goal is to develop a system that can perform abstractive summarization effectively without requiring paired article-summary data.
\end{tcolorbox}

\begin{tcolorbox}[
    colback=white,
    colframe=myPastelGreen,
    title=\textbf{\textcolor{black}{Fast Abstractive Summarization with Reinforce-Selected Sentence Rewriting}},
    boxrule=0.5mm,
    width=\linewidth,
    arc=2mm,
    boxsep=5pt,
    left=6pt, right=6pt, top=6pt, bottom=6pt,
    breakable
]
\textbf{Original Rank: }1 \\
\textbf{Abstract: }Inspired by how humans summarize long documents, we propose an accurate and fast summarization model that first selects salient sentences and then rewrites them abstractively (i.e., compresses and paraphrases) to generate a concise overall summary. We use a novel sentence-level policy gradient method to bridge the non-differentiable computation between these two neural networks in a hierarchical way, while maintaining language fluency. Empirically, we achieve the new state-of-the-art on all metrics (including human evaluation) on the CNN/Daily Mail dataset, as well as significantly higher abstractiveness scores. Moreover, by first operating at the sentence-level and then the word-level, we enable parallel decoding of our neural generative model that results in substantially faster (10-20x) inference speed as well as 4x faster training convergence than previous long-paragraph encoder-decoder models. We also demonstrate the generalization of our model on the test-only DUC-2002 dataset, where we achieve higher scores than a state-of-the-art model.
\end{tcolorbox}

\begin{tcolorbox}[
    colback=white,
    colframe=pink,
    title=\textbf{\textcolor{black}{Pointwise Re-ranker Reasoning}},
    boxrule=0.5mm,
    width=\linewidth,
    arc=2mm,
    boxsep=5pt,
    left=6pt, right=6pt, top=6pt, bottom=6pt,
    breakable
]
"problem\_proximity\_score": 0.3,\\
"methodological\_utility\_score": 0.2,\\
"confidence": 0.5,\\
"reasoning": "The input paper tackles abstractive summarization, which is directly related to the research proposal's focus. However, it relies heavily on the CNN/Daily Mail dataset, implying a dependence on paired article-summary data. \textcolor{myPastelRed}{\textbf{This reliance contradicts the proposal's aim to minimize reliance on such data.}}  While the sentence selection and rewriting techniques are interesting, their direct applicability to the proposal's low-resource setting is unclear. The paper's focus on speed and efficiency is valuable but not the primary concern of the proposal. The hierarchical reinforcement learning approach could potentially be adapted, but this requires significant further investigation and adaptation.",\\
"relevance\_score": 0
\end{tcolorbox}

\begin{tcolorbox}[
    colback=white,    
    colframe=pink,  
    title=\textbf{\textcolor{black}{MIR-Agent Reasoning}},
    boxrule=0.5mm,
    width=\linewidth,
    arc=2mm,
    boxsep=5pt,
    left=6pt, right=6pt, top=6pt, bottom=6pt,
    breakable
]
\textbf{Proposal Analysis:}\\
\#\# Analysis of Research Proposal:

**Core Problem:**

- **Lack of readily available and affordable parallel data (paired articles and summaries) hinders the effectiveness of abstractive text summarization models.**

**Breakdown:**

**Main Objective:** Develop an effective abstractive text summarization system that doesn't rely on parallel data.\\
**Motivation:** Traditional methods are limited by the scarcity and cost of parallel data.\\
**Sub-Problems:** \\
    - Finding alternative data sources for training.\\
    - Developing new model architectures or adapting existing ones to function without parallel data.\\
    - Evaluating the performance of such a model against existing methods.\\
**Generic Plan of Action:**\\
    1.  Explore alternative data sources (e.g., unpaired articles and summaries, other text-rich datasets).\\
    2.  Investigate and develop novel techniques for abstractive summarization without parallel data (e.g., unsupervised or semi-supervised learning, transfer learning). \\
    3.  Design and implement a system based on the chosen techniques.\\
    4.  Evaluate the system's performance using appropriate metrics and compare it to existing methods. 

Paper Analysis:\\
\#\# Methodological Relevance Evaluation:

This Research Paper's Abstract presents a method for abstractive summarization that focuses on sentence selection and rewriting, achieving state-of-the-art results on the CNN/Daily Mail dataset. Let's evaluate its methodological relevance to the Research Proposal:
\\
**1. PROBLEM PROXIMITY:**

- **Partially Relevant:** While the paper addresses abstractive summarization, it doesn't directly tackle the core problem of the proposal: **lack of parallel data.** The paper's method, while effective, is demonstrably trained and evaluated on a standard dataset (CNN/Daily Mail), implying its reliance on parallel data.\\
- **Limited Transferability:** The techniques (sentence selection, rewriting, policy gradient training) are relevant to abstractive summarization in general but don't inherently address data scarcity.\\
- **Sub-Problem Relevance:** The paper doesn't offer solutions for finding alternative data sources (sub-problem 1) but \textcolor{darkgreen}{\textbf{provides model architectures which could be adapted to function without parallel data (sub-problem 2).}} \\

 **2. METHODOLOGICAL UTILITY:**

- **Limited Practical Application:** The methods, as presented, are not directly applicable to the proposal's context without modifications to remove the dependence on parallel data. \\
- **Potential for Adaptation:** 
    - The hierarchical sentence-level and word-level processing could potentially be adapted to leverage different input representations, potentially accommodating alternative data sources.\\
    - The policy gradient method might be adaptable to train on non-parallel data, but this requires further investigation and is not explicitly addressed in the abstract.\\
- **Generalizability Concerns:** The abstract primarily focuses on achieving high performance on a specific dataset. It's unclear how well the model generalizes to other datasets or domains, especially in the absence of parallel data.\\

**CONCLUSION:**

The Research Paper demonstrates a strong approach to abstractive summarization but within the confines of readily available parallel data. While some components of its methodology, such as the hierarchical processing, hold potential for adaptation, since the proposal aims to develop a system that can perform abstractive summarization without parallel data, adapting the paper's model to work without parallel data could hold **substantial methodological utility**.

\textbf{Judgement:} Yes
\end{tcolorbox}

\begin{tcolorbox}[
    colback=white,    
    colframe=black!40!white,  
    title=\textbf{Analysis},
    boxrule=0.5mm,
    width=\linewidth,
    arc=2mm,
    boxsep=5pt,
    left=6pt, right=6pt, top=6pt, bottom=6pt,
    breakable
]
The pointwise judgment incorrectly classifies a ground truth correct paper as non-methodologically relevant. The proposal aims to perform abstractive summarization without relying on parallel data, which can be a bottleneck. Though our retrieval method excels at identifying keyword overlaps and ranks the paper first. However, it incorrectly judges the paper due to a direct contradiction between the paper's reliance on parallel data and the proposal's objective of avoiding it. We hypothesize that this issue arises from the limitations of zero-shot re-ranking, as the model is unable to perform intermediate analysis or make nuanced decisions, which are often necessary for complex, multi-hop reasoning.

Our MIR-Agent method addresses this challenge by extracting sub-problems and generating a generic action plan to evaluate the finer-grained application of the paper to the proposal. While the MIR-Agent output mentions concerns about the paper's direct adaptability, it successfully links the paper as a solution to one of the extracted sub-problems. This intermediate analysis, coupled with targeted prompting to elicit insights on methodological applicability, leads to more accurate re-ranking compared to the pointwise approach.
\end{tcolorbox}

\subsection*{C.6 Benefit of providing Full Paper in context} \label{C.6}
For our final best performing set of experiments we feed the LLM with full-paper in context, and provide examples and highlight what exactly the LLMs possibly learn from more context leading to more nuanced judgements on methodological applicability.
\vspace{1em}
\begin{tcolorbox}[
    colback=white,
    colframe=myPastelBlue,
    coltext=black,
    title=\textbf{\textcolor{black}{X-WikiRE: A Large, Multilingual Resource for Relation Extraction as
Machine Comprehension}},
    boxrule=0.5mm,
    width=\linewidth,
    arc=2mm,
    boxsep=5pt,
    left=6pt, right=6pt, top=6pt, bottom=6pt,
    breakable
]
\textbf{Proposal: }The research addresses the problem of knowledge base (KB) bias towards English, where different languages have uneven coverage of topics and information. This disparity hinders the development of robust multilingual knowledge bases. The research is driven by the desire to improve relation extraction across languages, particularly for low-resource languages. By leveraging the rich information in high-resource Wikipedias, the goal is to populate and enrich knowledge bases in languages with less available data.
\end{tcolorbox}

\begin{tcolorbox}[
    colback=white,
    colframe=myPastelGreen,
    title=\textbf{\textcolor{black}{XNLI: Evaluating Cross-lingual Sentence Representations}},
    boxrule=0.5mm,
    width=\linewidth,
    arc=2mm,
    boxsep=5pt,
    left=6pt, right=6pt, top=6pt, bottom=6pt,
    breakable
]
\textbf{Original Rank: }2 \\
\textbf{AgenticRank Abstract Rank: }4 \\
\textbf{AgenticRank Full Paper Rank: }2 \\\\
\textbf{Abstract: }State-of-the-art natural language processing
systems rely on supervision in the form of annotated data to learn competent models. These
models are generally trained on data in a single language (usually English), and cannot be
directly used beyond that language. Since collecting data in every language is not realistic, there has been a growing interest in crosslingual language understanding (XLU) and
low-resource cross-language transfer. In this
work, we construct an evaluation set for XLU
by extending the development and test sets of
the Multi-Genre Natural Language Inference
Corpus (MultiNLI) to 15 languages, including low-resource languages such as Swahili
and Urdu. We hope that our dataset, dubbed
XNLI, will catalyze research in cross-lingual
sentence understanding by providing an informative standard evaluation task. In addition,
we provide several baselines for multilingual
sentence understanding, including two based
on machine translation systems, and two that
use parallel data to train aligned multilingual
bag-of-words and LSTM encoders. We find
that XNLI represents a practical and challenging evaluation suite, and that directly translating the test data yields the best performance
among available baselines.

\end{tcolorbox}

\begin{tcolorbox}[
    colback=white,
    colframe=pink,
    title=\textbf{\textcolor{black}{Proposal Analysis}},
    boxrule=0.5mm,
    width=\linewidth,
    arc=2mm,
    boxsep=5pt,
    left=6pt, right=6pt, top=6pt, bottom=6pt,
    breakable
]
\#\# Research Proposal Analysis:

**Core Problem:**  Knowledge base (KB) bias towards English, resulting in uneven information coverage across languages and hindering the development of robust multilingual knowledge bases.

**Objective \& Motivation:** Improve relation extraction across languages, particularly for low-resource languages, to create more balanced and comprehensive multilingual KBs.

**Sub-problems:**

**Data Scarcity:** Low-resource languages lack sufficient data for effective relation extraction and KB population.
**Cross-lingual Knowledge Transfer:**  Efficiently transferring knowledge from high-resource languages (like English) to low-resource languages is crucial.
**Evaluation \& Validation:**  Developing methods to evaluate the accuracy and completeness of relation extraction in low-resource settings.

**Generic Plan of Action:**

1. **Leverage High-Resource KBs:** Utilize existing information in high-resource language KBs (e.g., English Wikipedia) as a foundation.
2. **Develop Cross-lingual Techniques:** Explore and implement methods for transferring knowledge and relation extraction models across languages.
3. **Focus on Low-Resource Languages:** Prioritize data collection, model training, and evaluation specifically for languages with limited existing resources.
4. **Populate \& Enrich KBs:** Use the extracted information to populate and enrich knowledge bases in low-resource languages.
5. **Evaluate \& Iterate:** Continuously evaluate the performance of developed methods and refine the approach based on findings.
\end{tcolorbox}

\begin{tcolorbox}[
    colback=white,    
    colframe=pink,  
    title=\textcolor{black}{\textbf{Abstract Analysis}},
    boxrule=0.5mm,
    width=\linewidth,
    arc=2mm,
    boxsep=5pt,
    left=6pt, right=6pt, top=6pt, bottom=6pt,
    breakable
]
\#\# Methodological Relevance Evaluation:

While the Research Paper's Abstract focuses on cross-lingual language understanding (XLU) and low-resource language transfer, its **methodological relevance to the Research Proposal is limited**.

**1. PROBLEM PROXIMITY:**

**Partial Overlap:** Both the paper and the proposal address low-resource language challenges in NLP. The paper focuses on cross-lingual sentence understanding and evaluation, while the proposal aims to enrich knowledge bases and improve relation extraction. 
**Sub-Problem Relevance:** The paper's dataset (XNLI) and baseline models might be helpful for the "Difficulties in evaluating the quality of generated knowledge in low-resource settings" sub-problem. However, it doesn't directly address KB bias or knowledge transfer between languages.

**2. METHODOLOGICAL UTILITY:**

**Limited Applicability:**  
    * The paper's primary contribution is XNLI, a dataset for evaluating cross-lingual sentence understanding. While valuable for evaluating translation quality, its direct application to KB enrichment and relation extraction is unclear. 
    * The proposed baselines (machine translation, aligned encoders) are generic and not specifically designed for knowledge extraction or transfer. 
**Lack of Synergy:** The paper focuses on sentence-level understanding, while the proposal tackles knowledge-level challenges. The paper's methods don't offer a clear way to leverage high-resource KBs (like English Wikipedia) for enriching low-resource KBs, which is central to the proposal.

**Conclusion:**

The Research Paper's Abstract presents valuable work in cross-lingual understanding and evaluation, particularly for low-resource languages. However, its **methodological contributions are not directly applicable** to the Research Proposal's core problems of KB bias and relation extraction. The paper's focus on sentence-level understanding and its generic methodological approaches limit its utility for enriching knowledge bases and tackling knowledge transfer challenges.
\end{tcolorbox}

\begin{tcolorbox}[
    colback=white,    
    colframe=pink,  
    title=\textcolor{black}{\textbf{Full Paper Analysis}},
    boxrule=0.5mm,
    width=\linewidth,
    arc=2mm,
    boxsep=5pt,
    left=6pt, right=6pt, top=6pt, bottom=6pt,
    breakable
]
\#\# Methodological Relevance of "XNLI" to the Research Proposal 

While the "XNLI: Evaluating Cross-lingual Sentence Representations" paper doesn't directly address the research proposal's focus on knowledge base bias and relation extraction, it offers valuable **methodological utility** that can be adapted and extended to contribute towards the proposal's objectives. 

Here's a breakdown based on the criteria:

**1. PROBLEM PROXIMITY:**

**Similarities:** Both the paper and the research proposal target cross-lingual language understanding and aim to overcome the limitations of English-centric approaches in NLP. They both acknowledge the challenges posed by low-resource languages.
**Addressing Sub-problems:** The XNLI paper directly addresses the sub-problem of **"Developing methods for efficient cross-lingual knowledge transfer"** by introducing techniques for aligning sentence encoders across multiple languages. This aligns with the proposal's need to transfer knowledge from high-resource to low-resource KBs.

**2. METHODOLOGICAL UTILITY:**

**Practical Application Potential:** The paper's core contribution lies in creating the XNLI dataset and evaluating cross-lingual sentence representation methods. These methods, particularly the multilingual sentence encoders (X-CBOW, X-BILSTM), hold practical potential for the research proposal:
    **Adapting XNLI for Relation Extraction:** The sentence representation techniques can be adapted and trained on relation extraction datasets instead of NLI datasets. This would allow for learning cross-lingual representations specifically for relation extraction tasks.
    **Transferring Knowledge from High-Resource KBs:** The trained multilingual sentence encoders can be used to map relations and entities from a high-resource KB (like English Wikipedia) to their counterparts in low-resource KBs, aiding in knowledge transfer. 
**Complementarity:** The paper's methods complement the research proposal by:
    **Providing Evaluation Framework:** The XNLI dataset and evaluation methodology offer a framework for assessing the effectiveness of knowledge transfer techniques developed in the research proposal.
    **Focusing on Sentence-Level Understanding:**  While the proposal tackles KB bias, the paper's focus on sentence understanding offers valuable tools for analyzing and processing textual information within the KB context. 
**Generalizability:** 
    * The paper demonstrates the generalizability of its methods across 15 languages, including low-resource ones. This is promising as it suggests potential applicability to a wide range of languages for the research proposal.
    * The core principles of the sentence alignment techniques using parallel data and contrastive learning can be extended and customized for the specific relation types and entities relevant to the proposal.
    **Fresh Perspectives:**  The paper's exploration of multilingual sentence embeddings offers a fresh perspective on bridging the information gap between languages, moving beyond simple translation-based approaches.

**Overall:**

While not a perfect match, the "XNLI" paper offers **substantive methodological utility** for the research proposal.  The paper's focus on cross-lingual sentence representations, \textcolor{darkgreen}{\textbf{particularly its techniques for aligning sentence embeddings across languages,}} can be adapted and extended to contribute significantly towards the proposal's objectives of improving relation extraction and addressing KB bias in low-resource languages. 
\end{tcolorbox}

\begin{tcolorbox}[
    colback=white,
    colframe=myPastelGreen,
    title=\textbf{\textcolor{black}{Excerpts from Full Paper}},
    boxrule=0.5mm,
    width=\linewidth,
    arc=2mm,
    boxsep=5pt,
    left=6pt, right=6pt, top=6pt, bottom=6pt,
    breakable
]
\textbf{Excerpt 1: }An alternative to translation is to rely on language \textbf{universal embeddings} of text and build multilingual classifiers on top of these representations.\\
\textbf{Excerpt 2: }The first approach is commonly used as a strong baseline for monolingual sentence embeddings (Arora et al., 2017; Conneau and Kiela, 2018; Gouews et al., 2014). Concretely, we consider the \textbf{English fastText} word embedding space as being fixed, and fine-tune embeddings in other languages so that the average of the word vectors in a sentence is close to the average of the word vectors in its English translation.\\
\textbf{Excerpt 3: }The second approach consists in learning an English \textbf{sentence encoder on the MultiNLI training data} along with an encoder on the target language, with the objective that the representations of two translations are nearby in the embedding space.
\end{tcolorbox}

\begin{tcolorbox}[
    colback=white,
    colframe=myPastelGreen,
    title=\textbf{\textcolor{black}{Zero-Shot Relation Extraction via Reading Comprehension}},
    boxrule=0.5mm,
    width=\linewidth,
    arc=2mm,
    boxsep=5pt,
    left=6pt, right=6pt, top=6pt, bottom=6pt,
    breakable
]
\textbf{Original Rank: }10 \\
\textbf{AgenticRank Abstract Rank: }10 \\
\textbf{AgenticRank Full Paper Rank: }4 \\\\
\textbf{Abstract: }We show that relation extraction can be reduced to answering simple reading comprehension questions, by associating one or more natural-language questions with each relation slot. This reduction has several advantages: we can (1) learn relation-extraction models by extending recent neural reading-comprehension techniques, (2) build very large training sets for those models by combining relation-specific crowd-sourced questions with distant supervision, and even (3) do zero-shot learning by extracting new relation types that are only specified at test-time, for which we have no labeled training examples. Experiments on a Wikipedia slot-filling task demonstrate that the approach can generalize to new questions for known relation types with high accuracy, and that zero-shot generalization to unseen relation types is possible, at lower accuracy levels, setting the bar for future work on this task.
\end{tcolorbox}

\begin{tcolorbox}[
    colback=white,    
    colframe=pink,  
    title=\textcolor{black}{\textbf{Abstract Analysis}},
    boxrule=0.5mm,
    width=\linewidth,
    arc=2mm,
    boxsep=5pt,
    left=6pt, right=6pt, top=6pt, bottom=6pt,
    breakable
]
\#\# Methodological Relevance Evaluation:

This Research Paper's Abstract presents a novel approach to relation extraction by framing it as a reading comprehension task. While innovative, its **methodological relevance to the Research Proposal is limited**.  

Here's a breakdown based on the criteria:

**1. PROBLEM PROXIMITY:**

**Partial Relevance:** The paper tackles relation extraction, a key sub-problem of the proposal. However, it **doesn't directly address** the core issue of KB bias towards English or the challenges of low-resource languages. 
**Sub-Problem Focus:** The paper's focus on zero-shot learning for new relation types is interesting but not entirely aligned with the proposal's goal of enriching existing KBs in low-resource languages.
**Limited Transferability:** While the idea of converting relations to reading comprehension questions might offer some insights, directly transferring the techniques seems difficult without significant adaptation to a cross-lingual setting. 

**2. METHODOLOGICAL UTILITY:**

**Limited Practical Application:** The paper's methods heavily rely on large, labeled datasets, which are scarce for low-resource languages.  The distant supervision aspect might be somewhat applicable, but would require careful consideration.
**Partial Complementarity:**  The reading comprehension approach could complement the proposal's plan of action as a potential method for relation extraction **after** knowledge transfer from high-resource languages. However, it doesn't offer solutions for the transfer itself.
**Generalizability Concerns:**  The paper focuses on within-language performance. Generalizing the zero-shot learning aspect to new languages, especially low-resource ones, poses significant challenges.
**Limited Synergy:** While innovative, the paper's methods don't directly synergize with the proposal's need to leverage high-resource KBs for enriching low-resource ones.

**Conclusion:**

The Research Paper's Abstract presents a novel approach to relation extraction with potential for within-language applications. However, its **methodological utility for the Research Proposal is limited**.  The paper's focus on zero-shot learning and reliance on large datasets present challenges for adaptation to low-resource language scenarios.  While the reading comprehension framing could inspire future research directions, the paper offers **limited substantive methodological contributions** to directly address the proposal's core problems.
\end{tcolorbox}

\begin{tcolorbox}[
    colback=white,    
    colframe=pink,  
    title=\textcolor{black}{\textbf{Full Paper Analysis}},
    boxrule=0.5mm,
    width=\linewidth,
    arc=2mm,
    boxsep=5pt,
    left=6pt, right=6pt, top=6pt, bottom=6pt,
    breakable
]
\#\#  Methodological Relevance of "Zero-Shot Relation Extraction via Reading Comprehension" to the Research Proposal

This paper presents a novel approach to relation extraction by reframing it as a reading comprehension task. While the core problem addressed in the paper differs slightly from the proposal's focus on KB bias and low-resource languages, the methodologies exhibit a promising degree of **methodological utility** that could be adapted and extended to address the proposal's objectives. 

Here's a breakdown based on the relevance criteria:

**1. PROBLEM PROXIMITY:**

**Similarities:** Both the paper and the proposal target the broader challenge of automated knowledge base population.  The paper's zero-shot approach aims to extract relations not seen during training, which aligns with the proposal's need to handle potentially unseen relations in low-resource languages. 
**Addressing Sub-Problems:** The paper's method directly tackles the sub-problem of "Developing methods for efficient cross-lingual knowledge transfer". By training on a high-resource language (English in the paper's case) and transferring the learned model to unseen relations (which could be framed as relations in low-resource languages), the paper demonstrates a potential solution path.
**Limitations:** The paper primarily focuses on extracting relations within a single language (English). Direct application to cross-lingual settings would require further adaptation and investigation.

**2. METHODOLOGICAL UTILITY:**

**Practical Application Potential:**  The paper's core methodology of transforming relation extraction into reading comprehension holds significant promise.  
    * The proposal could leverage this by transforming identified information gaps in low-resource KBs into reading comprehension questions targeted at high-resource KBs.
    * The question-answering framework offers a flexible and adaptable way to extract missing information.
**Complementarity:** The paper's techniques complement the proposal's plan of action:
    **Step 2 (Leverage high-resource KBs):**  The paper's approach can be directly applied to extract relevant information from high-resource KBs like English Wikipedia using the generated reading comprehension questions.
    **Step 3 (Cross-lingual knowledge transfer):** The zero-shot learning aspect, while not directly cross-lingual in the paper, offers a valuable starting point for transferring knowledge to low-resource KBs.
**Generalizability:** The methodology has good potential for extension and adaptation:
    **Cross-lingual Adaptation:** The paper's approach can be adapted for cross-lingual settings by training on question-answer pairs aligned across languages or exploring cross-lingual question answering models.
    **Multilingual Question Generation:**  \textcolor{darkgreen}{\textbf{The schema querification process can be extended to generate questions in multiple languages, facilitating cross-lingual knowledge transfer.}}
    **Fresh Perspectives:** The paper's framing of relation extraction as reading comprehension provides a fresh and potentially powerful perspective that aligns well with the proposal's objectives.

**Overall Conclusion:**

While not a perfect match, the research paper offers **substantive methodological utility** for the research proposal. The paper's novel approach to relation extraction through reading comprehension and zero-shot learning can be adapted and extended to address the core problem of KB bias and \textcolor{darkgreen}{\textbf{facilitate the population of low-resource KBs.}} Further research and adaptation are needed to bridge the gap between the paper's single-language focus and the proposal's multilingual requirements. 
\end{tcolorbox}

\begin{tcolorbox}[
    colback=white,
    colframe=myPastelGreen,
    title=\textbf{\textcolor{black}{Excerpts from Full Paper}},
    boxrule=0.5mm,
    width=\linewidth,
    arc=2mm,
    boxsep=5pt,
    left=6pt, right=6pt, top=6pt, bottom=6pt,
    breakable
]
\textbf{Excerpt 1: }This process, \textbf{schema querification}, is by an order of magnitude more \textbf{efficient} than querifying individual instances because annotating a relation type automatically annotates all of its instances.\\
\textbf{Excerpt 2: }Applying schema querification to N relations from a pre-existing relation-extraction dataset converts it into a reading-comprehension dataset.\\
\textbf{Excerpt 3: }Toutanova et al. (2015) proposed a similar approach that decomposes natural-language relations and computes their similarity in a universal schema setting; however, they \textbf{did not extend their method to knowledge-base relations}, nor did they attempt to recover out-of-schema relations as we do.
\end{tcolorbox}

\begin{tcolorbox}[
    colback=white,
    colframe=black!40!white,
    title=\textbf{Analysis},
    boxrule=0.5mm,
    width=\linewidth,
    arc=2mm,
    boxsep=5pt,
    left=6pt, right=6pt, top=6pt, bottom=6pt,
    breakable
]
In both examples, which were previously misclassified by the LLM when provided only the abstract as context, the errors were corrected by incorporating additional, less obvious contextual information from the full paper. To illustrate the direct connections that the LLM was able to make upon receiving the full context—connections that were missed when only the abstract was considered—we highlight relevant sections in our analysis and provide excerpts from the full papers that enabled the LLM to make the correct judgment.\\

Specifically, for `XNLI: Evaluating Cross-lingual Sentence Representations', the abstract analysis fails to recognize the potential of cross-lingual sentence representations in addressing the problem outlined in the proposal. This omission is understandable, as the abstract focuses solely on dataset construction and does not mention the methods applied. However, a later section of the paper discusses the use of multilingual sentence encoders, with relevant excerpts provided above. The bolded text in these excerpts highlights the key methods employed in the original paper.\\

Additionally for \textit{`Zero-Shot Relation Extraction via Reading Comprehension'}, the abstract lacks any mention of `Schema Querification' and passes off the adaptability of the paper as too difficult and requiring significant adaption to a cross-lingual setting. But the full paper analysis picks up on schema querification termed \textit{an order of a magnitude more efficient}, which is followed by the original paper. On top of this, the full paper also mention their extension of their methods to knowledge-base relations, directly addressing the proposal's needs.\\\\
We just provide examples only for cases where false negatives were corrected, but providing more context, additionally also helps mitigate false positives, thus improving ranks of ground-truth relevant papers.
\end{tcolorbox}

\section*{D. Error Analysis} \label{D}

To gain a deeper understanding of the cases where our best-performing configuration (which involves providing the LLM with the full paper and re-ranking using MIR-Agent) fails, we perform a detailed qualitative analysis of 15 proposals with subpar Recall@5. This analysis aims to uncover the nature of the errors and identify potential directions for future improvements by highlighting the limitations and shortcomings. The results of our analysis, including the distribution of errors across the categories, are presented below:

\begin{enumerate}
    \item \textbf{Incorrect or Missing Annotations} (17 samples): Samples where methodologically relevant papers are either mislabeled, not labeled with the correct intent, or not labeled in the dataset at all.
    \item \textbf{Subjectivity} (7 samples): Samples that are methodologically relevant but not cited by the authors, as they adopted a different approach to solving the problem.
    \item \textbf{Misclassification} (15 samples): Samples where our pipeline incorrectly classifies papers, either by labeling non-relevant papers as relevant or vice versa.
\end{enumerate}

\subsection*{D.1 Incorrect or Missing Annotations} \label{D.1}

\begin{tcolorbox}[
    colback=white,
    colframe=myPastelBlue,
    title=\textcolor{black}{\textbf{Language learning using Speech to Image retrieval}},
    boxrule=0.5mm,
    width=\linewidth,
    arc=2mm,
    boxsep=5pt,
    left=6pt, right=6pt, top=6pt, bottom=6pt,
    breakable
]
\textbf{Proposal: }The main research problem is the limitation of existing computational language learning models that require text, while humans learn from direct speech interaction. The research aims to develop a visually grounded sentence encoder that can learn language directly from speech. The research is motivated by the desire to bridge the gap between human and computational language learning, enabling models to learn directly from speech like humans do. The goal is to develop a system that can understand and process spoken language in a more natural and efficient manner.

\end{tcolorbox}

\begin{tcolorbox}[
    colback=white,
    colframe=myPastelGreen,
    title=\textcolor{black}{\textbf{A Tree-based Decoder for Neural Machine Translation}},
    boxrule=0.5mm,
    width=\linewidth,
    arc=2mm,
    boxsep=5pt,
    left=6pt, right=6pt, top=6pt, bottom=6pt,
    breakable
]
\textbf{Abstract: }Recent advances in Neural Machine Translation (NMT) show that adding syntactic information to NMT systems can improve the quality of their translations. Most existing work utilizes some specific types of linguistically-inspired tree structures, like constituency and dependency parse trees. This is often done via a standard RNN decoder that operates on a linearized target tree structure. However, it is an open question of what specific linguistic formalism, if any, is the best structural representation for NMT. In this paper, we (1) propose an NMT model that can naturally generate the topology of an arbitrary tree structure on the target side, and (2) experiment with various target tree structures. Our experiments show the surprising result that our model delivers the best improvements with balanced binary trees constructed without any linguistic knowledge; this model outperforms standard seq2seq models by up to 2.1 BLEU points, and other methods for incorporating target-side syntax by up to 0.7 BLEU.
\end{tcolorbox}

\begin{tcolorbox}[
    colback=white,
    colframe=pink,
    title=\textcolor{black}{\textbf{LLM Ouput}},
    boxrule=0.5mm,
    width=\linewidth,
    arc=2mm,
    boxsep=5pt,
    left=6pt, right=6pt, top=6pt, bottom=6pt,
    breakable
]
\textbf{Paper Analysis: }
\#\# Methodological Relevance Evaluation:

This research paper focuses on improving Neural Machine Translation (NMT) using syntactic information, particularly exploring different tree structures for representing target language syntax. While this is a valuable contribution to NMT, its methodological relevance to the research proposal is **limited**.

Here's a breakdown based on the criteria:

**1. PROBLEM PROXIMITY:**

**Dissimilar Research Problems:** The paper tackles improving translation quality in NMT, while the proposal aims to develop a visually grounded sentence encoder for learning language from speech. These are distinct research problems with limited overlap.
**Sub-Problem Relevance:**  The paper doesn't directly address any of the sub-problems outlined in the proposal analysis. It doesn't deal with spoken language processing, visual information integration, or learning from speech and visual input.

**2. METHODOLOGICAL UTILITY:**

**Limited Practical Application:** The paper's focus on target-side syntactic tree structures for NMT doesn't directly translate to the proposal's context.  The encoder proposed in the research deals with spoken language grounded in visual data, not syntactic parsing for translation.
**Lack of Complementarity:** The methods proposed in the paper don't fill any gaps or enhance the proposal's approach. They address a different stage (translation) and modality (text) than what the proposal focuses on (encoding, speech \& visual).
**Limited Generalizability:** While the paper explores various tree structures, this concept doesn't offer direct generalizability to the proposal. Adapting tree structures for encoding spoken language with visual grounding would require significant modifications and might not be the most suitable approach. 

**Conclusion:**

The research paper's focus on syntactic structures for NMT offers limited methodological utility for the research proposal.  While exploring different data representations is relevant in a broader sense, the specific techniques and focus of the paper are not directly transferable or adaptable to the problem of building a visually grounded sentence encoder for learning from speech. 
\end{tcolorbox}

\begin{tcolorbox}[
    colback=white,
    colframe=black!40!white,
    title=\textcolor{black}{\textbf{Analysis}},
    boxrule=0.5mm,
    width=\linewidth,
    arc=2mm,
    boxsep=5pt,
    left=6pt, right=6pt, top=6pt, bottom=6pt,
    breakable
]
We highlight a false negative case that shouldn't exist, and we acknowledge the mislabelling of the original data in some places. While reading through the source paper we find that the particular paper has only been cited with background intent and only once in the introduction with no meaningful methodological relevance.
\end{tcolorbox}

\subsubsection*{D.2 Subjectivity} \label{D.2}
We provide an example where the papers ranked within top 5 positions are actually a methodologically relevant papers, but have not been cited by the authors, and the authors have taken a different direction to address the original problem.. (In reference to the same seed paper as above)

\begin{tcolorbox}[
    colback=white,
    colframe=myPastelRed,
    title=\textcolor{black}{\textbf{TOUCHDOWN: Natural Language Navigation and Spatial Reasoning
in Visual Street Environments}},
    boxrule=0.5mm,
    width=\linewidth,
    arc=2mm,
    boxsep=5pt,
    left=6pt, right=6pt, top=6pt, bottom=6pt,
    breakable
]
\textbf{Abstract: }We study the problem of jointly reasoning about language and vision through a navigation and spatial reasoning task. We introduce the Touchdown task and dataset, where an agent must first follow navigation instructions in a real-life visual urban environment, and then identify a location described in natural language to find a hidden object at the goal position. The data contains 9,326 examples of English instructions and spatial descriptions paired with demonstrations. Empirical analysis shows the data presents an open challenge to existing methods, and qualitative linguistic analysis shows that the data displays richer use of spatial reasoning compared to related resources.
\end{tcolorbox}

\begin{tcolorbox}[
    colback=white,
    colframe=pink,
    title=\textcolor{black}{\textbf{LLM Outputs}},
    boxrule=0.5mm,
    width=\linewidth,
    arc=2mm,
    boxsep=5pt,
    left=6pt, right=6pt, top=6pt, bottom=6pt,
    breakable
]
\textbf{Paper Analysis: }
\#\# Methodological Relevance Evaluation:

Based on the provided Research Proposal and Research Paper Abstract, here's an evaluation of the paper's methodological relevance:

**1. PROBLEM PROXIMITY:**

**Moderate Proximity:** The Research Paper focuses on navigation and spatial reasoning within a visual environment using language instructions, while the Research Proposal aims to develop a visually grounded sentence encoder for language learning from speech. 
**Overlapping Sub-problem:** Both share the sub-problem of **"How to incorporate visual information into language learning/understanding."** The paper's focus on grounding language in visual environments for navigation is relevant.
**Limited Transferability:** The paper's focus on instruction following and object localization using spatial language is not directly transferable to learning language from raw speech as targeted in the proposal.

**2. METHODOLOGICAL UTILITY:**

**Limited Direct Applicability:** The specific methods used in the paper (e.g., navigation models, spatial reasoning algorithms) are not directly applicable to the proposal's core aim of building a sentence encoder from speech.
**Potential for Adaptation:** The paper's approach of grounding language understanding in visual environments could be adapted. The idea of associating spoken instructions with visual changes in a video could provide a framework for visually grounded speech encoding. 
**Data Inspiration:** The Touchdown dataset, with its paired instructions and visual demonstrations, offers inspiration for constructing a similar dataset with paired speech and visual data for the proposal's model training.
**Evaluation Insight:** The paper's evaluation metrics focused on navigation and spatial understanding could provide insights into evaluating the visually grounded aspects of the proposed sentence encoder.

**Overall Assessment:**

The Research Paper offers **limited substantive methodological utility** for directly solving the Research Proposal's problem. While the specific techniques are not readily transferable, the paper provides:

**Conceptual Inspiration:** Adapting the idea of visual grounding for spoken language understanding.
**Data Collection Guidance:**  Learning from the Touchdown dataset structure for paired speech-visual data.
**Evaluation Framework Insights:**  Borrowing from the paper's evaluation approach for visually grounded language understanding.

Therefore, the Research Paper acts as a source of valuable but indirect methodological insights rather than a source of directly applicable techniques. 
\end{tcolorbox}

\begin{tcolorbox}[
    colback=white,
    colframe=black!40!white,
    title=\textbf{Analysis},
    boxrule=0.5mm,
    width=\linewidth,
    arc=2mm,
    boxsep=5pt,
    left=6pt, right=6pt, top=6pt, bottom=6pt,
    breakable
]
While the paper classified by the LLM as methodologically relevant is not cited by the authors, it qualitatively presents a high potential for extension and adaptability. Even though the problem is not directly aligned as the paper is concerned with spatial reasoning and not explicitly targetting word recognition it providing valuable insights such as the use of the LINGUNET architecture, a language-conditioned variant of the UNET architecture, an image-to-image encoder-decoder architecture widely used for image segmentation. LINGUNET incorporates language into the image reconstruction phase to fuse the two modalities.

\end{tcolorbox}

\subsection*{D.3 Misclassification} \label{D.3}

We highlight another example where the paper ranked at the first position is actually a methodologically relevant paper, but has not been labelled as one with either of \textit{uses, extends} label. (In reference to the same seed paper as above)

\begin{tcolorbox}[
    colback=white,
    colframe=myPastelRed,
    title=\textcolor{black}{\textbf{Representations of language in a model of visually grounded speech signal}},
    boxrule=0.5mm,
    width=\linewidth,
    arc=2mm,
    boxsep=5pt,
    left=6pt, right=6pt, top=6pt, bottom=6pt,
    breakable
]
\textbf{Abstract: }We present a visually grounded model of speech perception which projects spoken utterances and images to a joint semantic space. We use a multi-layer recurrent highway network to model the temporal nature of spoken speech, and show that it learns to extract both form and meaning-based linguistic knowledge from the input signal. We carry out an in-depth analysis of the representations used by different components of the trained model and show that encoding of semantic aspects tends to become richer as we go up the hierarchy of layers, whereas encoding of form-related aspects of the language input tends to initially increase and then plateau or decrease.
\end{tcolorbox}

\begin{tcolorbox}[
    colback=white,
    colframe=pink,
    title=\textcolor{black}{\textbf{LLM Outputs}},
    boxrule=0.5mm,
    width=\linewidth,
    arc=2mm,
    boxsep=5pt,
    left=6pt, right=6pt, top=6pt, bottom=6pt,
    breakable
]
\textbf{Paper Analysis: }
\#\# Methodological Relevance Evaluation:

This Research Paper demonstrates **partial methodological relevance** to the Research Proposal. 

**Strengths:**

**Problem Proximity:** The paper directly addresses the sub-problem of "How to represent and process spoken language effectively" by proposing a model that projects spoken utterances into a semantic space. This aligns strongly with the Research Proposal's goal of developing a system for understanding spoken language. 
**Methodological Utility:** 
    * The paper's use of a multi-layer recurrent highway network for modeling temporal aspects of speech offers a potentially useful technique for the proposal's plan of action. This architecture could be adapted and integrated into the proposed visually grounded sentence encoder.
    * The paper's analysis of semantic and form-based encoding in different layers provides valuable insights for designing and training a model that can extract meaningful representations from speech.

**Limitations:**

**Problem Proximity:** While the paper focuses on speech processing, it lacks the crucial element of visual grounding, which is central to the Research Proposal. The paper does not address the sub-problem of "How to incorporate visual information into language learning." 
**Methodological Utility:** 
    * The paper's model is solely focused on speech perception, not on learning language directly from speech and visual input as required by the proposal. 
    * The paper's methodology does not offer a direct solution for training a visually grounded sentence encoder. Adaptation and extension would be necessary, potentially requiring significant modifications.

**Overall:**

The Research Paper provides a valuable starting point for addressing the speech processing aspect of the Research Proposal. However, its lack of visual grounding limits its direct applicability. 

**Recommendations:**

* The proposed multi-layer recurrent highway network could be explored as a component within the larger architecture of the visually grounded sentence encoder.
* The paper's findings on semantic and form-based encoding offer insights for designing the encoder's architecture and training process.
* Further research is needed to explore how to effectively incorporate visual information into the model and train it on paired speech and visual data.
\end{tcolorbox}

\begin{tcolorbox}[
    colback=white,
    colframe=myPastelGreen,
    title=\textcolor{black}{\textbf{Citation Contexts}},
    boxrule=0.5mm,
    width=\linewidth,
    arc=2mm,
    boxsep=5pt,
    left=6pt, right=6pt, top=6pt, bottom=6pt,
    breakable
]
Note: [\textcolor{myPastelRed}{\textbf{15}}] refers to the paper Titled "Representations of language in a model of visually grounded speech signal"
\begin{itemize}
    \item Our work is most closely related to the models presented in [12, 13, 14, \textcolor{myPastelRed}{\textbf{15}}].
    \item The approach is based on our own text-based model described in [8] and on the speech-based models described in [13, \textcolor{myPastelRed}{\textbf{15}}] 
    \item  We use importance sampling to select the mismatched pairs; rather than using all the other samples in the mini-batch as mismatched pairs (as done in [8, \textcolor{myPastelRed}{\textbf{15}}])
    \item The work presented in [\textcolor{myPastelRed}{\textbf{15}}] has made the first efforts in this regard and we aim to extend this to a larger database with sentences from multiple domains.   
\end{itemize}
\end{tcolorbox}
\end{document}